\documentclass[journal,compsoc]{IEEEtran}
\usepackage{amsmath,amsfonts}
\usepackage{algorithmic}
\usepackage{algorithm}
\usepackage{array}
\usepackage[caption=false,font=normalsize,labelfont=sf,textfont=sf]{subfig}
\usepackage{textcomp}
\usepackage{stfloats}
\usepackage{url}
\usepackage{verbatim}
\usepackage{graphicx}
\ifCLASSOPTIONcompsoc
  \usepackage[nocompress]{cite}
\else
  \usepackage{cite}
\fi

\usepackage[hashEnumerators,hybrid]{markdown}
\usepackage{graphicx}
\usepackage{amsmath}
\usepackage{amssymb}
\usepackage{booktabs}
\usepackage{xcolor}
\usepackage[utf8x]{inputenc}
\usepackage{colortbl}
\usepackage{placeins}
\usepackage{multirow}
\usepackage{tabularx}
\usepackage{pgf}
\usepackage{tikz}
\usepackage{pifont}
\usepackage[final,tracking=true,kerning=true,spacing=true,factor=1100,stretch=10,shrink=10]{microtype}

\usepackage[pagebackref,breaklinks,colorlinks]{hyperref}

\newcommand*{\cm}{\checkmark}
\newcommand*{\g}{\color{gray}}
\newcommand{\spm}[1]{\tiny{$\,\pm$#1}}

\DeclareMathOperator*{\argmax}{arg\,max}

\newcolumntype{Y}{>{\centering\arraybackslash}X}

\begin{document}

\title{
Domain Adaptive and Generalizable Network Architectures and Training Strategies for Semantic Image Segmentation
}

\author{Lukas Hoyer, Dengxin Dai, and Luc Van Gool
        % <-this % stops a space
\IEEEcompsocitemizethanks{
\IEEEcompsocthanksitem Lukas Hoyer is with the ETH Zurich, Switzerland.
\protect\\ E-mail: lhoyer@vision.ee.ethz.ch
\IEEEcompsocthanksitem Dengxin Dai is with the Huawei Zurich Research Center, Switzerland.
\protect\\ E-mail: dengxin.dai@huawei.com
\IEEEcompsocthanksitem Luc Van Gool is with the ETH Zurich, Switzerland, the KU Leuven, Belgium, and the INSAIT Sofia, Bulgaria. E-mail: vangool@vision.ee.ethz.ch
}
}% <-this % stops a space

\IEEEtitleabstractindextext{%
\begin{abstract}
Unsupervised domain adaptation (UDA) and domain generalization (DG) enable machine learning models trained on a source domain to perform well on unlabeled or even unseen target domains. As previous UDA\&DG semantic segmentation methods are mostly based on outdated networks, we benchmark more recent architectures, reveal the potential of Transformers, and design the DAFormer network tailored for UDA\&DG. It is enabled by three training strategies to avoid overfitting to the source domain: While (1) Rare Class Sampling mitigates the bias toward common source domain classes, (2) a Thing-Class ImageNet Feature Distance and (3) a learning rate warmup promote feature transfer from ImageNet pretraining. As UDA\&DG are usually GPU memory intensive, most previous methods downscale or crop images. However, low-resolution predictions often fail to preserve fine details while models trained with cropped images fall short in capturing long-range, domain-robust context information. Therefore, we propose HRDA, a multi-resolution framework for UDA\&DG, that combines the strengths of small high-resolution crops to preserve fine segmentation details and large low-resolution crops to capture long-range context dependencies with a learned scale attention. DAFormer and HRDA significantly improve the state-of-the-art UDA\&DG by more than 10 mIoU on 5 different benchmarks. The implementation is available at \texttt{\url{https://github.com/lhoyer/HRDA}}.
\end{abstract}

}

\maketitle

\section{Introduction}
\label{sec:introduction}

\begin{figure}
    \centering
    \includegraphics[width=0.95\linewidth]{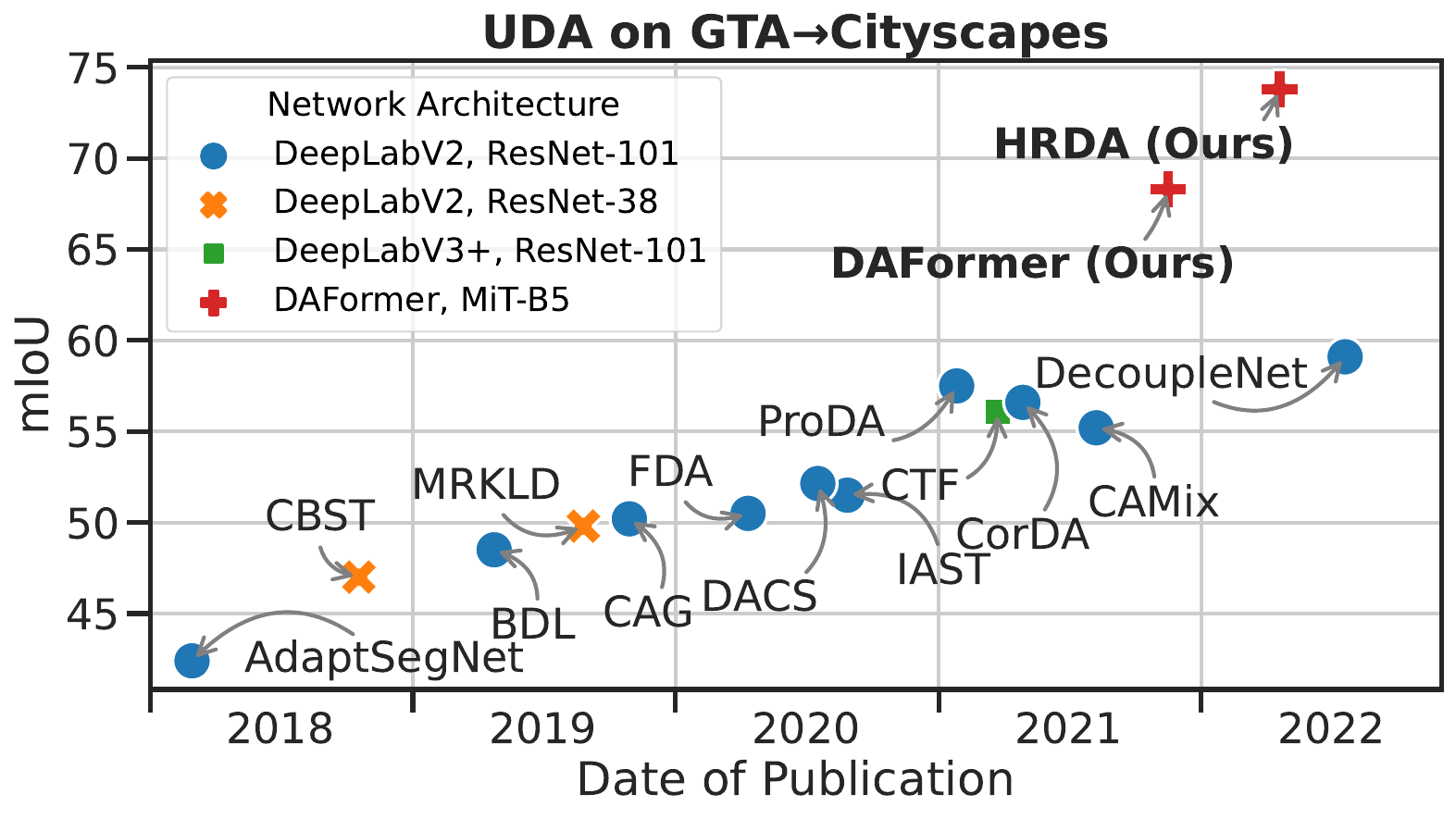}
    \includegraphics[width=0.95\linewidth]{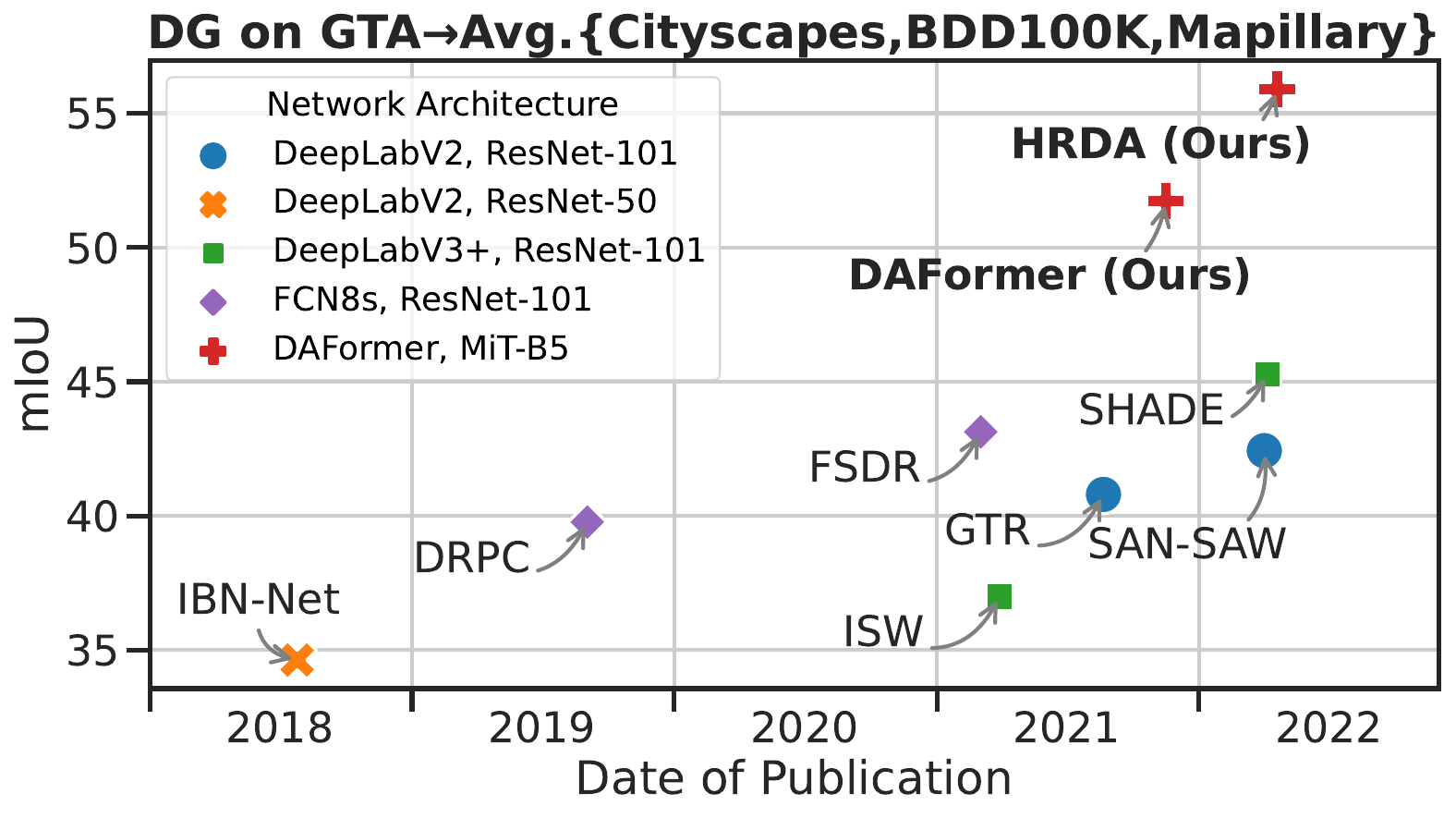}
    \caption{Progress of UDA and DG over time. Most previous UDA and DG methods are evaluated using the outdated DeepLab architecture. We rethink the design of the network architecture as well as its training strategies for UDA and DG. The proposed DAFormer and HRDA significantly outperform previous methods for both UDA and DG.}
    \label{fig:uda_over_time}
\end{figure}

\IEEEPARstart{I}{n} the last few years, neural networks have achieved overwhelming performance on many computer vision tasks. However, they require a large amount of annotated data to be trained properly. For semantic image segmentation, annotations are particularly costly as every pixel has to be labeled. For instance, it takes 1.5 hours to annotate a single image of Cityscapes~\cite{cordts2016cityscapes} while it is even 3.3 hours for adverse weather conditions~\cite{sarkadis2021acdc}.
To avoid the annotation effort of the target domain, where the network should be deployed, the network can instead be trained on a source dataset with existing or cheaper annotations such as automatically labeled synthetic data~\cite{ros2016synthia, richter2016playing}. However, neural networks are usually sensitive to domain shifts. Therefore, domain generalization (DG) methods utilize specific training strategies to improve the domain robustness so that the network better generalizes from the source to unseen target domains. In unsupervised domain adaptation (UDA), the network is additionally adapted to the known target domain using unlabeled target images.
In the following, we introduce our enhanced UDA\&DG framework. It consists of a network architecture specifically tailored for domain robustness, several training strategies to avoid overfitting to the source domain, and a multi-crop/-resolution framework to enable context-aware high-resolution UDA and DG.

\noindent\textbf{Domain-Robust Network Architecture:}
Previous UDA and DG methods mostly evaluated their contributions using network architectures, which are outdated in the area of supervised semantic segmentation. For instance, the commonly used DeepLabV2~\cite{chen2017deeplab} only achieves a supervised performance of 65 mIoU~\cite{tsai2018learning} on Cityscapes while more recent networks reach up to 85 mIoU~\cite{yuan2020object, tao2020hierarchical}.
Due to the large performance gap, it stands to question whether using outdated network architectures can limit the overall UDA and DG performance and misguide the benchmark progress. To answer this question, this work studies the influence of the network architecture for UDA and DG, compiles a more sophisticated architecture, and successfully applies it to UDA and DG with a few simple, yet crucial training strategies.
Naively using a more powerful network architecture for UDA and DG might be suboptimal as it can be more prone to overfitting to the source domain.
Based on a study of different semantic segmentation architectures evaluated for UDA and DG, we compile \emph{DAFormer}, a network architecture tailored for domain adaptation and generalization (Sec.~\ref{sec:methods_architecture}). It is based on recent Transformers~\cite{dosovitskiy2020image, xie2021segformer}, which have been shown to be more robust than the predominant CNNs~\cite{bhojanapalli2021understanding}. We combine them with a context-aware multi-level feature fusion, which further enhances domain robustness. 

\noindent\textbf{Domain-Robust Training Strategies:}
Since more complex and capable architectures are more prone to overfitting to the source domain, we introduce three training strategies to UDA and DG to address these issues (Sec.~\ref{sec:methods_training_strategies}).
First, we propose \emph{Rare Class Sampling (RCS)} to consider the long-tail distribution of the source domain, which hinders the learning of domain-robust representations of rare classes for DG and UDA. Particularly in UDA, this is critical due to the confirmation bias of self-training toward common classes. By frequently sampling images with rare classes, RCS enables the network to learn domain-robust decision boundaries more stably. In UDA, this additionally improves the quality of pseudo-labels and reduces the confirmation bias.
Second, we propose a \emph{Thing-Class ImageNet Feature Distance (FD)}, which distills knowledge from diverse and expressive ImageNet features to regularize the source training. This is particularly helpful as the source domain is limited to only a few instances of certain classes (low diversity), which have a different appearance than the target domain (domain shift). Without FD this would result in learning less expressive and source-domain-specific features, which hinder domain generalization and adaptation. As ImageNet features were trained for thing-classes, we restrict the FD to regions of the image that are labeled as a thing-class.
And third, we introduce learning rate warmup~\cite{goyal2017accurate} newly to UDA\&DG. By using a small learning rate during early training, the learning process is stabilized and features from ImageNet pretraining can be better transferred to semantic segmentation.

\noindent\textbf{Domain-Robust Multi-Resolution Framework:}
DG and UDA methods are usually more GPU memory intensive than supervised learning as their training strategies often require images from multiple domains/styles, additional networks (e.g. teacher model or domain discriminator), and additional losses, which consume significant GPU memory.
Therefore, most UDA semantic segmentation methods (e.g.~\cite{tsai2018learning, tranheden2021dacs, araslanov2021self, wang2021domain}) follow the convention of downscaling images to satisfy GPU memory constraints.
However, predictions from low-resolution (LR) inputs often fail to recognize small objects such as distant traffic lights and to preserve fine segmentation details such as limbs of distant pedestrians. Alternatively, memory consumption can be reduced by training with random crops of high-resolution (HR) images. While this allows learning small objects and preserving segmentation details, it limits the learned long-range context dependencies to the crop size. This is a crucial disadvantage for DG and UDA as context information and scene layout are often domain-robust (e.g. rider on bicycle or sidewalk at the side of road)~\cite{huang2020contextual, yang2021context}. Further, HR inputs can be disadvantageous compared to LR inputs when generalizing/adapting large stuff-regions such as close sidewalks to another domain (see Sec.~\ref{sec:exp_resolution_crop_size}).
At HR, these regions often contain too detailed, domain-specific features (e.g. detailed sidewalk texture), which can be detrimental to domain robustness. LR inputs `hide away' these features, while still providing sufficient details to recognize large regions across domains.

To effectively combine the strength of both approaches, we propose \emph{HRDA}, a novel multi-resolution framework for DG and UDA semantic segmentation.
First, HRDA uses a large LR \emph{context crop} to adapt large objects without confusion from domain-specific HR textures and to learn long-range context dependencies as we assume that HR details are not crucial for long-range dependencies.
Second, HRDA uses a small HR \emph{detail crop} from the region within the context crop to adapt small objects and to preserve segmentation details as we assume that long-range context information play only a subordinate role in learning segmentation details. In that way, the GPU memory consumption is significantly reduced while still preserving the advantages of a large crop size and a high resolution. 
Given that the importance of the LR context crop vs. the HR detail crop depends on the content of the image, HRDA fuses both using an input-dependent scale attention. It learns to decide how trustworthy the LR and the HR predictions are in every image region.

\noindent\textbf{Contributions:}
This work is based on our previous papers DAFormer~\cite{hoyer2021daformer} (published at CVPR'22) and HRDA~\cite{hoyer2022hrda} (published at ECCV'22). Both have shown a great potential for synthetic-to-real UDA, set a new performance standard for this task, and are being actively used by the community. In this work, we unify both methods  and extend them beyond synthetic-to-real UDA to facilitate a more comprehensive view on domain robustness. In particular, this work newly contributes the following aspects:

\begin{itemize}
    \item While DAFormer and HRDA were originally designed for UDA, we newly extended both to DG. In a comprehensive ablation study, the proposed components display significant improvements in DG, demonstrating their effectiveness beyond UDA. Overall, DAFormer and HRDA significantly improve the state-of-the-art DG performance. We believe that the extension of DAFormer and HRDA to DG is an important step for the research community as DG semantic segmentation has received less attention than UDA even though it is of significant practical relevance as the exact target domain is often not known in advance in many applications.
    \item Further, we additionally study the capabilities of DAFormer and HRDA on day-to-nighttime and clear-to-adverse-weather UDA as these scenarios have a significant practical relevance in autonomous driving. Also in these scenarios, DAFormer and HRDA achieve strong performance improvements over previous works.
    \item We unite DAFormer and HRDA in a coherent method for domain adaptation and generalization. Alongside the paper, we provide a comprehensive open-source framework for UDA\&DG with all the provided benchmarks to facilitate future research on domain-adaptive and -generalizable semantic segmentation. DAFormer and HRDA can be trained in one stage on a single consumer GPU (16 hours on an RTX~2080~Ti 16 hours for DAFormer and 32 hours on a Titan RTX for HRDA), which simplifies its usage compared to previous methods such as ProDA~\cite{zhang2021prototypical}, which requires training multiple stages on four V100 GPUs for several days.
    \item Overall, HRDA and DAFormer represent a major advance in UDA and DG. They outperform previous UDA and DG methods by a large margin of more than +10 mIoU across 5 diverse benchmarks (see Fig.~\ref{fig:uda_over_time}), demonstrating that the network architecture, appropriate training strategies, and multi-resolution training play an important role for UDA and DG.
    Specifically, they outperform previous methods by +16.3 on GTA$\rightarrow$Cityscapes, by +10.3 on Synthia$\rightarrow$Cityscapes, by +11.6 on Cityscapes$\to$DarkZurich, by +18.0 on Cityscapes$\to$ACDC, and by +10.6 for DG on GTA$\to$Avg.\{Cityscapes,BDD100K,Mapillary\}.
\end{itemize}

\section{Related Work}
\label{sec:related_work}

\subsection{Semantic Image Segmentation}

Since the breakthrough of neural network in semantic segmentation~\cite{long2015fully}, they were improved in various aspects such as integrating context~\cite{chen2018encoder,zhang2018context,hoyer2019grid,yuan2020object}, attention mechanisms~\cite{wang2018non,fu2019dual}, and Transformer backbones~\cite{vaswani2017attention, xie2021segformer}.

Several architectures~\cite{chen2017deeplab,chen2018encoder,lin2019zigzagnet} utilize intermediate features with different scales, which are generated from a single scale input, to aggregate context information. 
Furthermore, multi-scale input inference, where predictions from scaled versions of an image are combined via average or max pooling, is often used to obtain better results~\cite{chen2018encoder,xie2021segformer}. However, the naive pooling is independent of the image content, which can lead to suboptimal results. Therefore, Chen et al.~\cite{chen2016attention} and Yang et al.~\cite{yang2018attention} segment multi-scale inputs and learn an attention-weighted sum of the predictions. Tao et al.~\cite{tao2020hierarchical} further propose a hierarchical attention that is agnostic to the number of scales during inference. 

For image classification, CNNs were shown to be sensitive to distribution shifts such as image corruptions~\cite{hendrycks2018benchmarking} or domain shifts~\cite{hendrycks2021many}. Recent works~\cite{bhojanapalli2021understanding, naseer2021intriguing} show that Transformers are more robust than CNNs with respect to these properties. While CNNs focus on textures~\cite{geirhos2018imagenet}, Transformers put more importance on the object shape~\cite{bhojanapalli2021understanding, naseer2021intriguing}, which is more similar to human vision~\cite{geirhos2018imagenet}.
For semantic segmentation, ASPP~\cite{chen2018encoder} and skip connections were reported to increase the robustness~\cite{kamann2021benchmarking}. Further, Xie~et al.~\cite{xie2021segformer} showed that their Transformer-based architecture improves the robustness.
However, the influence of recent network architectures on the DG and UDA performance of semantic segmentation has not been systematically studied yet.

\subsection{Domain Generalization (DG)}
\label{sec:related_work_domain_generalization}

DG aims to learn domain invariant representations based on the source domain(s) that generalize well to unseen  domains. DG methods for semantic segmentation generally follow two directions: On the one hand, methods based on instance normalization and whitening~\cite{pan2018two, choi2021robustnet} explicitly use these transformations to remove domain-specific information from the features. On the other hand, domain randomization methods aim to broaden the source domain by augmenting the data with different styles \cite{yue2019domain, peng2021global, zhao2022style, zhongadversarial, zhao2022stylea}.
Concurrently to this work, \cite{zhao2022stylea} integrates our DAFormer architecture and independently confirms its merits for DG.

\subsection{Unsupervised Domain Adaptation (UDA)}
\label{sec:related_work_uda}

UDA methods aim to adapt a network to an unlabeled target domain. Most approaches can be grouped into adversarial training and self-training. Adversarial training methods aim to align the distributions of source and target domain at input~\cite{hoffman2018cycada, gong2021dlow}, feature~\cite{hoffman2016fcns, tsai2018learning}, or output~\cite{tsai2018learning, vu2019advent} level in a GAN framework~\cite{goodfellow2014generative}.
In self-training, the network is adapted using high-confidence pseudo-labels of the target domain. To regularize the training and to avoid pseudo-label drift, approaches such as confidence thresholding~\cite{zou2018unsupervised}, curriculum adaptation~\cite{dai2020curriculum,dai2018dark}, pseudo-label prototypes~\cite{zhang2021prototypical}, and consistency regularization based on data augmentation~\cite{araslanov2021self,hoyer2022mic}, different context~\cite{zhou2021context}, or domain-mixup~\cite{tranheden2021dacs, hoyer2021improving} have been used.

Datasets often follow a long-tail distribution, which biases models toward common classes~\cite{wang2017learning}. In UDA, this problem has been approached by loss re-weighting~\cite{zou2018unsupervised} and class-balanced sampling for image classification~\cite{prabhu2021sentry}. We extend the latter from classification to semantic segmentation and propose Rare Class Sampling (RCS), which addresses the co-occurrence of rare and common classes in a single semantic segmentation sample. Further, we demonstrate that RCS is particularly effective to train Transformers for UDA.

Li~et al.~\cite{li2017learning} have shown that knowledge distillation from an old task can act as a regularizer for a new task. It can improve semi-supervised learning~\cite{hoyer2021three} and adversarial UDA~\cite{chen2018road}. We apply this idea to self-training UDA, show its particular benefits for Transformers, and improve it by restricting the knowledge distillation to image regions with thing-classes as ImageNet mostly labels thing-classes.

The use of semantic segmentation networks with multi-scale \emph{features} is quite common in UDA due to the common use of DeepLabV2~\cite{chen2017deeplab}. However, these features are generated from a single-scale input.
While some works apply multi-scale average pooling for inference~\cite{araslanov2021self, wang2020classes} or enforce scale consistency~\cite{subhani2020learning,iqbal2020mlsl} of low-resolution inputs, they fall short in learning an input-adaptive multi-scale fusion.
To the best of our knowledge, HRDA is the first work to learn a multi-resolution \emph{input} fusion for UDA and DG semantic segmentation.
For that purpose, we newly extend scale attention~\cite{chen2016attention,tao2020hierarchical} to UDA and DG and reveal its significance for domain robustness by improving the adaptation/generalization process across different object scales.
Further, we propose fusing nested crops with different scales and sizes, which successfully overcomes the pressing issue of limited GPU memory for multi-resolution UDA and DG.

\section{Methods}
\label{sec:methods}

After defining the problem setup of DG and UDA (Sec.~\ref{sec:methods_problem_setup}), we outline the baseline approaches for DG and UDA (Sec.~\ref{sec:methods_dg} and \ref{sec:methods_self_training}). Subsequently, we discuss the DAFormer network architecture and the training strategies for improving domain robustness (Sec.~\ref{sec:methods_training_strategies}) including Rare Class Sampling, the Thing-Class ImageNet Feature Distance (FD), and the Learning Rate Warmup. Finally, the HRDA high/multi-resolution training for DG and UDA is presented (Sec.~\ref{sec:methods_hrda}).

\subsection{Problem Setup}
\label{sec:methods_problem_setup}

In DG, a neural network $f_\theta$ is trained on a labeled source domain $\mathcal{S} = \{(x^{S}_m, y^{S}_m)\}_{m=1}^{N_S}$ with images $x^{S}_m \in \mathbb{R}^{H_S \times W_S \times 3}$ and their corresponding annotation $y^{S}_m \in \{0,1\}^{H_S \times W_S \times K}$. To estimate the domain generalization capability of $f_\theta$, it is evaluated on an unseen target domain $\mathcal{T}$, which usually shares the same label space of $K$ classes. In UDA, the training process has additional access to \emph{unlabeled} target domain images $\mathcal{T}= \{x^{T}_m\}_{m=1}^{N_T}$ with $x^{T}_m \in \mathbb{R}^{H_T \times W_T \times 3}$. As the following definitions refer to the same source/target sample, we will drop index $m$ to avoid convolution.

As only source labels are available, the supervised categorical cross-entropy loss can only be calculated for the source predictions $\hat{y}^S = f_\theta(x^S)$
\begin{align}
    \mathcal{L}^S &= \mathcal{L}_\mathit{ce}(\hat{y}^S, y^S, 1)\,,\\
    \mathcal{L}_\mathit{ce}(\hat{y}, y, q) &= -\sum_{i=1}^{H} \sum_{j=1}^{W} \sum_{k=1}^K q_{ij} y_{ijk} \log \hat{y}_{ijk}\,.
\end{align}

\subsection{Style Diversification for DG}
\label{sec:methods_dg}

To improve the domain generalization of the model, typical approaches include removing style information from the features~\cite{pan2018two, choi2021robustnet} and diversifying the source domain training data with additional styles\cite{yue2019domain, peng2021global, zhao2022style}. We opt for the latter group as it enables benchmarking different network architectures without modifying the underlying architectures. In particular, we start with a simple DG baseline that uses naive photometric distortion with random brightness, contrast, saturation, and hue augmentation to diversify the style of the source domains. For more advanced DG, we follow SHADE~\cite{zhao2022style} and diversify the style of encoder features via AdaIN~\cite{huang2017arbitrary} by interpolating new styles from a set of basis styles. The basis styles are obtained by farthest point sampling over the different styles within the source domain. Further, a style consistency loss minimizes the Jensen-Shannon Divergence between the posterior probability original source image and its stylized variant. For further details, we refer the readers to SHADE~\cite{zhao2022style}.

\begin{figure*}
    \centering
    \includegraphics[width=\linewidth]{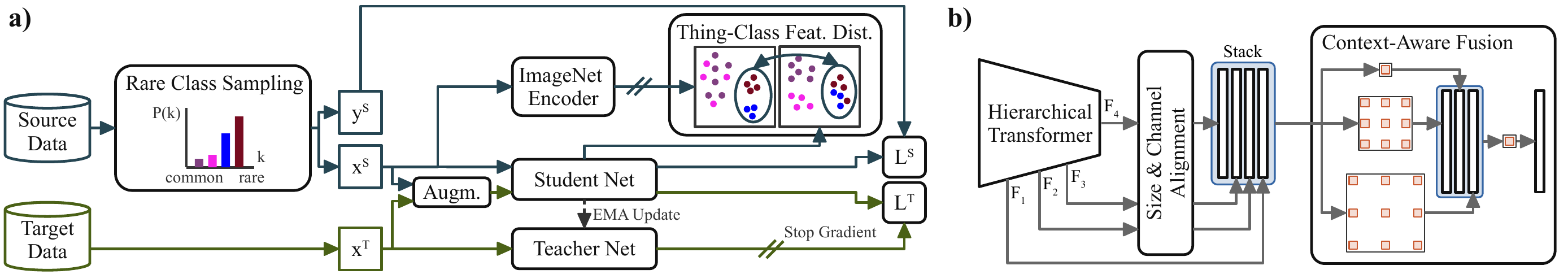}
    \caption{Overview of our DG and UDA framework with Rare Class Sampling, Thing-Class Feature Distance, and DAFormer network. The green components are only used in UDA as no target data is available in DG.}
    \label{fig:daformer_overview}
\end{figure*}

\subsection{Self-Training for UDA}
\label{sec:methods_self_training}

To adapt the model to the target domain, UDA methods incorporate an additional loss for the target domain $\mathcal{L}^T$.
The target loss can be defined according to the used training strategies such as adversarial training~\cite{tsai2018learning,wang2020classes} or self-training~\cite{zou2018unsupervised,tranheden2021dacs,zhang2021prototypical}. 
We use self-training as adversarial training is known to be less stable and is currently outperformed by self-training methods~\cite{tranheden2021dacs, zhang2021prototypical}.
In self-training, the model $f_\theta$ is iteratively adapted to the target domain using pseudo-labels $p^T$ and their quality estimates $q^T$
\begin{equation}
    \mathcal{L}^T = \mathcal{L}_\mathit{ce}(\hat{y}^T, p^T, q^T)\,.
\end{equation}
The pseudo-labels are produced by a teacher network $g_\phi$
\begin{equation}
    p_{ijk}^T = [k = \argmax_{k'} g_\phi(x^T)_{ijk'}]\,,
\end{equation}
where $[\cdot]$ denotes the Iverson bracket.
Note that no gradients are backpropagated into the teacher.
The pseudo-labels can be generated either online~\cite{araslanov2021self,tranheden2021dacs} or offline~\cite{zou2018unsupervised, zou2019confidence}.
We opted for online self-training due to its less complex setup with only one training stage. This is important as we compare and ablate various network architectures.
In online self-training, the teacher model $g_\phi$ is updated after each training step $t$ with the exponentially moving average of the weights of $f_\theta$
\begin{equation}
    \phi_{t+1} \leftarrow \alpha \phi_t + (1 - \alpha) \theta_t\,,
\end{equation}
implementing a temporal ensemble to stabilize pseudo-labels~\cite{tarvainen2017mean}.
For the quality estimate $q^T$, we follow DACS~\cite{tranheden2021dacs} and use the ratio of pixels exceeding a threshold $\tau$ of the maximum softmax probability of $g_\phi(x^T)$.
Self-training is particularly efficient if the student $f_\theta$ is trained on augmented target data, while the teacher $g_\phi$ generates the pseudo-labels using non-augmented target data~\cite{tarvainen2017mean, tranheden2021dacs, araslanov2021self}.
Here, we use color jitter, blur, and ClassMix~\cite{olsson2020classmix} as augmentation.

\subsection{DAFormer Network Architecture}
\label{sec:methods_architecture}

In this work, we compile a network architecture that is specifically tailored for DG and UDA to not just achieve good supervised performance but also provide good domain generalization and adaptation capabilities.

For the encoder, we aim for a powerful yet robust network architecture. 
We hypothesize that general robustness is an important property to achieve good domain generalization/adaptation performance as it fosters the learning of domain-invariant features.
Based on recent findings~\cite{bhojanapalli2021understanding, naseer2021intriguing} as well as an architecture comparison for DG and UDA, which we will present in Sec.~\ref{sec:exp_comparison_networks}, Transformers~\cite{vaswani2017attention} are a good choice for DG and UDA as they fulfill these criteria. Although the self-attention from Transformers and the convolution both perform a weighted sum, their weights are computed differently: in CNNs, the weights are learned during training but fixed during testing; in the self-attention mechanism, the weights are dynamically computed based on the similarity or affinity between every pair of tokens. As a consequence, the self-similarity operation in the self-attention mechanism provides modeling means that are potentially more adaptive and general than convolution operations.
In particular, we follow the design of Mix Transformers (MiT)~\cite{xie2021segformer}, which are tailored for semantic segmentation. The image is divided into small patches of a size of $4{\times}4$  (instead of $16{\times}16$ as in ViT~\cite{dosovitskiy2020image}) to preserve details for semantic segmentation. To cope with the high feature resolution, sequence reduction is used in the self-attention blocks. The transformer encoder is designed to produce multi-level feature maps $F_i \in \mathbb{R}^{\frac{H}{2^{i+1}} \times \frac{W}{2^{i+1}} \times C_i}$. The downsampling of the feature maps is implemented by overlapping patch merging~\cite{xie2021segformer} to preserve local continuity.

Previous works on semantic segmentation with Transformers usually exploit only local information for the decoder~\cite{xie2021segformer, zheng2021rethinking}.
In contrast, we propose to utilize additional context information in the decoder
as this has been shown to increase the general robustness of semantic segmentation~\cite{kamann2021benchmarking}, a helpful property for DG and UDA. Instead of just considering the context information of the bottleneck features~\cite{chen2017deeplab, chen2018encoder}, DAFormer uses the context across features from different encoder levels as the additional earlier features provide valuable low-level concepts for semantic segmentation at a high resolution, which can also provide important context information.
The architecture of the DAFormer decoder is shown in Fig.~\ref{fig:daformer_overview}\,b. Before the feature fusion, we embed each $F_i$ to the same number of channels $C_{e}$ by a $1{\times}1$ convolution, bilinearly upsample the features to the size of $F_1$, and concatenate them. For the context-aware feature fusion, we use multiple parallel $3{\times}3$ depthwise separable convolutions with different dilation rates and a $1{\times}1$ convolution to fuse them, similar to ASPP~\cite{chen2018encoder} but without global average pooling. In contrast to the original use of ASPP~\cite{chen2018encoder}, we do not only apply it to the bottleneck features $F_4$ but use it to fuse all stacked multi-level features. Depthwise separable convolutions have the advantage that they have a lower number of parameters than regular convolutions, which can reduce overfitting to the source domain.

\subsection{Domain-Robust Training Strategies}
\label{sec:methods_training_strategies}

One challenge of training a more capable network for DG and UDA is overfitting to the source domain. To circumvent this issue, we introduce three strategies to stabilize and regularize DG and UDA: Rare Class Sampling, Thing-Class ImageNet Feature Distance, and learning rate warmup. The enhanced DG and UDA pipeline is shown in Fig.~\ref{fig:daformer_overview}\,a.

\subsubsection{Rare Class Sampling (RCS)}
\label{sec:methods_rare_class_sampling}

Even though the more capable DAFormer is able to achieve better performance on difficult classes than other architectures,
we observed that the UDA performance for classes that are rare in the source dataset varies significantly over different runs.
Depending on the random seed of the data sampling order, these classes are learned at different iterations of the training or sometimes not at all as we will show in Sec.~\ref{sec:exp_rcs}.
The later a certain class is learned during the training, the worse is its performance at the end of the training. 
We hypothesize that if relevant samples containing rare classes only appear late in the training due to randomness, the network only starts to learn them later, and more importantly, it is highly possible that the network has already learned a strong bias toward common classes making it difficult to `re-learn' new concepts with very few samples in a domain-robust way. This is further reinforced as the bias is confirmed by self-training with the teacher network.

To address this, we propose Rare Class Sampling (RCS). It samples images with rare classes from the source domain more often to learn them 
better and earlier.
The frequency $\omega_k$ of each class $k$ in the source dataset can be calculated based on the number of pixels with class $k$
\begin{equation}
    \omega_k = \frac{\sum_{m=1}^{N_S} \sum_{i=1}^{H} \sum_{j=1}^{W} [y^S_{mijk}]}{N_S \cdot H \cdot W}\,.
\end{equation}
The sampling probability $P(k)$ of a certain class $k$ is defined as a function of its frequency $\omega_k$
\begin{equation}
    P(k) = \frac{e^{(1-\omega_k) / T}}{\sum_{k'=1}^K e^{(1-\omega_{k'}) / T}}\,.
    \label{eq:P_k}
\end{equation}
Therefore, classes with a smaller frequency will have a higher sampling probability. The temperature $T$ controls the smoothness of the distribution. A higher $T$ leads to a more uniform distribution, a lower $T$ to a stronger focus on rare classes with a small $\omega_k$.
For each source sample, a class is sampled from the probability distribution $k \sim P$ and an image is sampled from the subset of data containing this class $x_S \sim \text{uniform}(\mathcal{X}_{S,k})$.
Eq.~\ref{eq:P_k} allows to over-sample images containing rare classes ($P(k) \geq 1 / K$ if $\omega_k$ is small). 
As a rare class (small $\omega_k$) usually co-occurs with multiple common classes (large $\omega_k$) in a single image, it is beneficial to sample rare classes more often than common classes ($P(k_\textit{rare}) > P(k_\textit{common})$) to get closer to a balance of the re-sampled classes.
For example, the common class road co-occurs with rare classes such as bus, train, or motorcycle and is therefore already covered when sampling images with these rare classes.
When decreasing $T$, more pixels of classes with small $\omega_k$ are sampled but also fewer pixels of classes with medium $\omega_k$. The temperature $T$ is chosen to reach a balance of the number of re-sampled pixels of classes with small and medium $\omega_k$ by maximizing the number of re-sampled pixels of the class with the least.

\subsubsection{Thing-Class ImageNet Feature Distance (FD)}
\label{sec:methods_feature_distance}
Commonly, the semantic segmentation model $f_\theta$ is initialized with weights from ImageNet classification to start with meaningful generic features. Given that ImageNet contains images from some relevant high-level semantic classes, which DG and UDA methods often struggle to distinguish such as train or bus, we hypothesize that the ImageNet features can provide useful guidance beyond the usual pretraining. In particular, we observe that the DAFormer network is able to segment some classes at the beginning of the training but forgets them later (see Sec.~\ref{sec:exp_fd}).
Therefore, we assume that useful features from ImageNet pretraining are corrupted by $\mathcal{L}^S$ and the model overfits to the source data.

To prevent this issue, we regularize the model based on the feature distance $d$ of the bottleneck features $F_\theta$ of the semantic segmentation model $f_\theta$ and the bottleneck features $F_\mathit{ImageNet}$ of the ImageNet model
\begin{equation}
    d^S_{ij} = ||F_{ImageNet}(x^S)_{ij} - F_\theta(x^S)_{ij}||_2\,.
\end{equation}
However, the ImageNet model is mostly trained on thing-classes (objects with a well-defined shape such as car or zebra) instead of stuff-classes (amorphous background regions such as road or sky). Therefore, we calculate the FD loss only for image regions containing thing-classes $\mathcal{K}_\mathit{things}$ described by the binary mask $M^\mathit{S,things}$
\begin{equation}
    \mathcal{L}^S_\mathit{FD} = \frac{\sum_{i=1}^{H_F} \sum_{j=1}^{W_F} d^S_{ij} \cdot M^\mathit{S,things}_{ij}}{\sum_{i=1}^{H_F} \sum_{j=1}^{W_F} M^\mathit{S,things}_{ij}}\,,
\end{equation}
This mask is obtained from the downscaled label $y^{S, \mathit{small}}$
\begin{equation}
    M^\mathit{S,things}_{ij} = \sum_{k'=1}^K y^{S,\mathit{small}}_{ijk'} \cdot [k' \in \mathcal{K}_\mathit{things}]\,.
\end{equation}
To downsample the label to the bottleneck feature size, average pooling with a patch size $\frac{H}{H_F} {\times} \frac{W}{W_F}$ is applied to each class channel and a class is kept when it exceeds the ratio $r$.
This ensures that only bottleneck feature pixels containing a dominant thing-class are considered for the feature distance.

The loss functions for DG and UDA are supplemented with the weighted FD loss $\lambda_\mathit{FD} \mathcal{L}^S_\mathit{FD}$.

\subsubsection{Learning Rate Warmup}
\label{sec:methods_warmup}

Warming up the learning rate~\cite{goyal2017accurate} at the beginning of the training has successfully been used to train both CNNs~\cite{he2016deep} and Transformers~\cite{vaswani2017attention, dosovitskiy2020image} as it improves in-domain network generalization by avoiding that a large adaptive learning rate variance distorts the gradient distribution at the beginning of the training~\cite{liu2019variance}. We newly introduce learning rate warmup to DG and UDA. We posit that this is particularly important for domain robustness as distorting the features from ImageNet pretraining would deprive the network of useful guidance toward the real domain. During the warmup period $t_\mathit{warm}$, the learning rate at iteration $t$ is set $\eta_t = \eta_\mathit{base} \cdot t / t_\mathit{warm}$.

\subsection{Domain-Robust Multi-Resolution Training (HRDA)}
\label{sec:methods_hrda}

Further, we propose a multi-resolution framework for DG and UDA as small objects and segmentation details are easier to generalize and adapt with high-resolution (HR) inputs, while large stuff regions are easier to generalize and adapt with low-resolution (LR) inputs.
To maintain a managable GPU memory footprint in DG and UDA, we combine a large LR context crop to learn long-range context dependencies and a small HR detail crop to preserve segmentation details (Sec.~\ref{sec:context_detail_crop}). The strengths of both LR context and HR detail crop are combined by fusing their predictions with a learned scale attention (Sec.~\ref{sec:multi_resolution_uda}).
For a robust pseudo-labels in UDA, we further utilize overlapping slide inference to fuse predictions with different contexts (Sec.~\ref{sec:overlapping_slide_inference}).

\subsubsection{Context and Detail Crop}
\label{sec:context_detail_crop}

\begin{figure*}
\centering

\begin{minipage}[t]{0.78\linewidth}
\vspace{0pt}
\includegraphics[width=\linewidth]{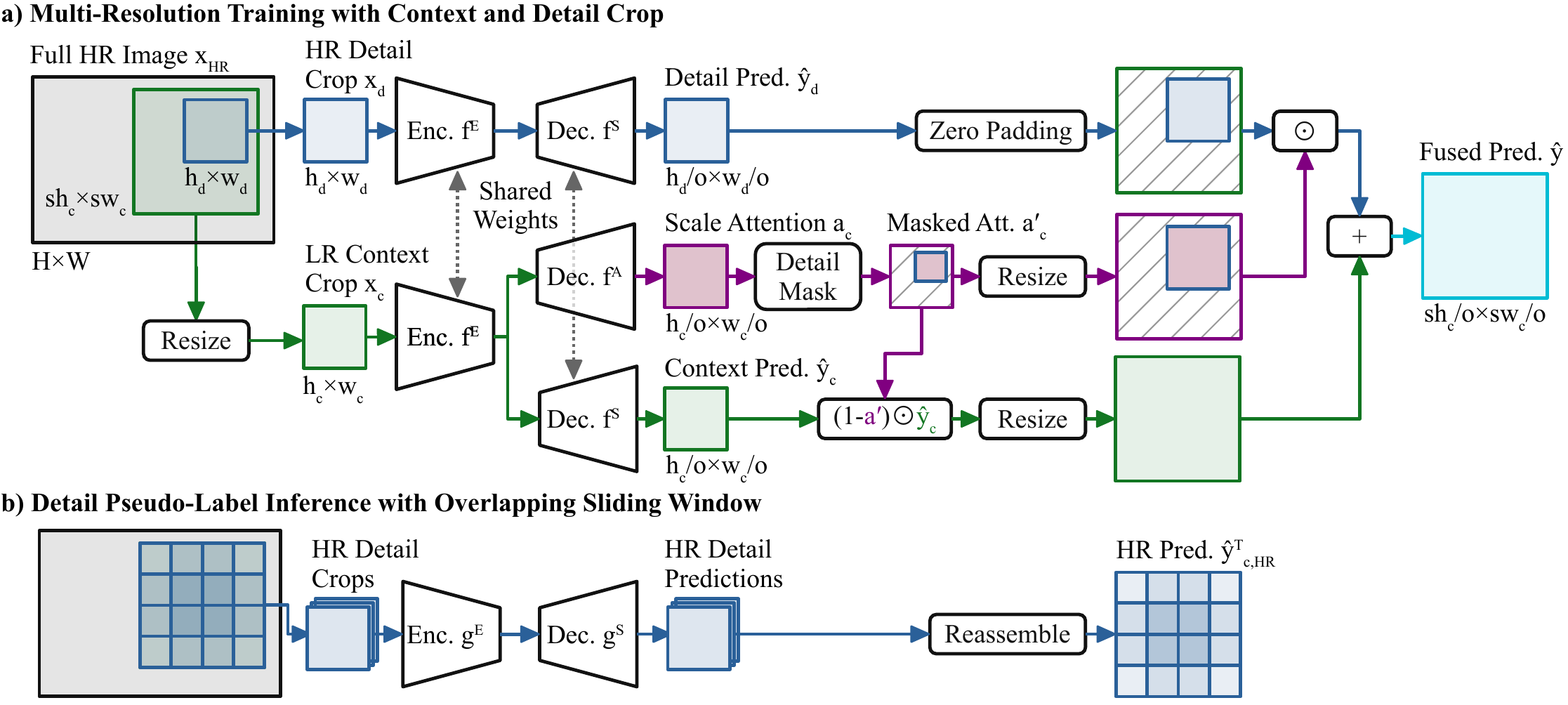}
\end{minipage}\hfill
\begin{minipage}[t]{0.2\linewidth}
\vspace{0pt}
\caption{(a) Multi-resolution training with low-resolution (LR) context and high-resolution (HR) detail crop. The prediction of the detail crop is fused into the context prediction within the region where it was cropped from by a learned scale attention. (b) For pseudo-label generation in UDA, multiple detail crops are generated using overlapping slide inference to cover the entire context crop. The pseudo-label is fused from HR pred. $\hat{y}_\mathit{c,HR}^T$ and LR pred. $\hat{y}_\mathit{c}^T$ with the full attention $a_c^T$ similar to (a) (see Sec.~\ref{sec:overlapping_slide_inference}). 
}
\label{fig:architecture}
\end{minipage}

\end{figure*}

Due to GPU memory constraints, it is often not feasible to train state-of-the-art DG and UDA methods with full-sized high-/multi-resolution inputs as images from multiple styles/domains, additional networks, and additional losses are often required for DG and UDA training. Therefore, most previous DG and UDA methods train with downscaled or cropped images. 
However, on the one side, LR inputs limit the ability to recognize small objects and produce fine segmentation borders. On the other side, random cropping restricts learning context-aware semantic segmentation, especially for long-range dependencies and scene layout, which might be critical for DG and UDA as context relations are often domain-invariant (e.g. car on road, rider on bicycle)~\cite{huang2020contextual, yang2021context}.
To train with long-range context and high resolution, we propose to combine different crop sizes for different resolutions, i.e. a large LR context crop and a small HR detail crop (see Fig.~\ref{fig:architecture}\,a). The purpose of the context crop is to provide a large crop to learn long-range context relations. The purpose of detail crop is to focus on HR to recognize small objects and produce fine segmentation details, which does not necessarily require far-away context information.

The context crop $x_c \in \mathbb{R}^{h_c \times w_c \times 3}$ is obtained by cropping from the original HR image $x_\mathit{HR} \in \mathbb{R}^{H \times W \times 3}$ and bilinear downsampling $\zeta(\cdot, \cdot)$ by the downscale factor $s \geq 1$
\begin{equation}
    x_{\mathit{c,HR}} = x_\mathit{HR}[b_{c,1} {:} b_{c,2}, b_{c,3} {:} b_{c,4}]\,, \hspace{2mm} x_c = \zeta(x_{c,\mathit{HR}}, 1/s)
    \label{eq:x_c}
\end{equation}
The crop bounding box $b_c$ is randomly sampled from a discrete uniform distribution within the image size while ensuring that the coordinates can be divided by $u=s \cdot o$ with $o \geq 1$ denoting the output stride of the segmentation network to ensure exact alignment in the later fusion process:
\begin{alignat}{3}
    b_{c,1} &\sim \mathcal{U}\{0, (H-sh_c) / u\} \cdot u\,,\quad &&b_{c,2} = b_{c,1} + sh_c\,,\\ 
    b_{c,3} &\sim \mathcal{U}\{0, (W-sw_c) / u\} \cdot u\,,\quad &&b_{c,4} = b_{c,3} + sw_c\,.\nonumber
\end{alignat}
The detail crop $x_d \in \mathbb{R}^{h_d \times w_d \times 3}$ is randomly cropped from within the context crop region to enable the fusion of context and detail predictions in the later process:
\begin{alignat}{3}
    x_d &= x_\mathit{c,HR}[b_{d,1} : b_{d,2}, b_{d,3} : b_{d,4}] &&\,,\label{eq:x_d}\\
    b_{d,1} &\sim \mathcal{U}\{0, (sh_c-h_d) / u\} \cdot u\,,\quad &&b_{d,2} &&= b_{d,1} + h_d\,,\\ 
    b_{d,3} &\sim \mathcal{U}\{0, (sw_c-w_d) / u\} \cdot u\,,\quad &&b_{d,4} &&= b_{d,3} + w_d\,.\nonumber
\end{alignat}
In this work, we use context and detail crops of the same dimension, i.e. $h_c{=}h_d$ and $w_c{=}w_d$, to balance the required resources for both crops and provide a good trade-off between context-aware and detailed predictions. The context downscale factor is $s{=}2$ following previous UDA methods~\cite{tsai2018learning,tranheden2021dacs}. So, the context crop covers 4 times more content at half the resolution compared to the detail crop.

Using a feature encoder $f^E$ and a semantic decoder $f^S$, the context semantic segmentation $\hat{y}_c = f^S(f^E(x_c)) \in \mathbb{R}^{\frac{h_c}{o} \times \frac{w_c}{o} \times K}$ and the detail semantic segmentation $\hat{y}_d = f^S(f^E(x_d)) \in \mathbb{R}^{\frac{h_d}{o} \times \frac{w_d}{o} \times K}$ are predicted. 
The network is shared for both HR and LR inputs to increase its robustness against different resolutions and to save GPU memory.

\subsubsection{Multi-Resolution Fusion}
\label{sec:multi_resolution_uda}

While the HR detail crop is well-suited to segment small objects such as poles or distant pedestrians, it lacks the ability to capture long-range dependencies, which is disadvantageous in segmenting large stuff regions such as large regions of sidewalk. The opposite is the case for the LR context crop. Therefore, we fuse the predictions from both crops using a learned scale attention~\cite{chen2016attention} to predict in which image regions to trust predictions from context and detail crop. 
Additionally, the scale attention can improve domain robustness by predicting objects at the better-suited scale
as the appearance of an object should have a resolution high enough to be discriminative but not too high to avoid that the network overfits to domain-specific detailed textures.

The scale attention decoder $f^A$ learns to predict the scale attention $a_c = \sigma(f^A(f^E(x_c))) \in [0,1]^{\frac{h_c}{o} \times \frac{w_c}{o} \times K}$ to weigh the trustworthiness of LR context and HR detail predictions. The sigmoid function $\sigma$ ensures a weight in $[0, 1]$.
The attention is predicted from the context crop as it has a better grasp on the scene layout (larger context).
As the predictions are smaller than the inputs due to the output stride $o$, the crop coordinates are scaled accordingly in the following steps. 
Outside of the detail crop $c_d$, the attention is set to 0 as there is no detail prediction available resulting in $a'_c = \mathit{DetailMask}(a_c, b_d) \in \mathbb{R}^{\frac{h_c}{o} \times \frac{w_c}{o}}$.
The detail crop is aligned with the (upsampled) context crop by zero-padding: $\hat{y}_d' = \mathit{ZeroPad}(\hat{y}_d, b_d) \in \mathbb{R}^{\frac{sh_c}{o} \times \frac{sw_c}{o}}$.
The predictions from multiple scales are fused using the attention-weighted sum
\begin{equation}
    \hat{y}_\mathit{c,F} = \zeta((1 - a'_c) \odot \hat{y}_c, s) + \zeta(a'_c, s) \odot \hat{y}_d'\,.
    \label{eq:scale_fusion}
\end{equation}
The encoder, segmentation head, and attention head are trained with the fused and the detail prediction
\begin{equation}
    \mathcal{L}_\mathit{HRDA}^S = (1 - \lambda_d) \mathcal{L}_\mathit{ce}(\hat{y}_{c,F}^S, y_{c,\mathit{HR}}^S, 1) + \lambda_d \mathcal{L}_\mathit{ce}(\hat{y}_{d}^S, y_{d}^S, 1)\,,
\end{equation}
where the ground truth $y_\mathit{c,HR}^S$/$y_d^S$ is cropped according to Eq.~\ref{eq:x_c}/\ref{eq:x_d}. Additionally supervising the detail crop is helpful to learn more robust features for HR inputs even though the attention might favor the context crop in that region. An additional loss for $\hat{y}_c$ is not necessary as it is already directly supervised in regions without detail crop.
The target loss $\mathcal{L}_\mathit{HRDA}^T$ is adapted accordingly
\begin{equation}
    \mathcal{L}_{\mathit{HRDA}}^{T} = 
    (1 - \lambda_d) \mathcal{L}_\mathit{ce}(\hat{y}_{c,F}^{T}, p_{c,F}^{T}, q_{c,F}^{T})
    + \lambda_d \mathcal{L}_\mathit{ce}(\hat{y}_{d}^{T}, p_{d}^{T}, q_{d}^{T})\,.
    \label{eq:L_T,HRDA}
\end{equation}
For pseudo-label prediction in UDA, we also utilize multi-resolution fusion (see Sec.~\ref{sec:overlapping_slide_inference}). Therefore, when predicting pseudo-labels, the scale attention focuses on the better-suited resolution (e.g. HR for small objects). As the pseudo-labels are further used to train the model also with the worse-suited resolution (e.g. LR for small objects), it improves the robustness for both small and large objects. 

\subsubsection{Pseudo-Labels with Overlapping Sliding Window}
\label{sec:overlapping_slide_inference}

For self-training with Eq.~\ref{eq:L_T,HRDA}, it is necessary to generate a high-quality HRDA pseudo-label $p_\mathit{c,F}^T$ for the context crop $x_\mathit{c,HR}^T$. The underlying HRDA prediction $\hat{y}_\mathit{c,F}^T$ is fused from the LR prediction $\hat{y}_c^T$ and HR prediction $\hat{y}_\mathit{c,HR}^T$ using the full scale attention $a_c^T$ similar to Eq.~\ref{eq:scale_fusion} \begin{equation}
    \hat{y}_\mathit{c,F}^T = \zeta((1 - a_c^T) \odot \hat{y}_c^T, s) + \zeta(a_c^T, s) \odot \hat{y}_\mathit{c,HR}^T\,.
\end{equation}
Therefore, the HR prediction $\hat{y}_\mathit{c,HR}^T$ is necessary for the entire context crop $x_\mathit{c,HR}^T$ instead of just the detail crop $x_d$.
We infer the HR prediction $\hat{y}_\mathit{c,HR}^T$ using a sliding window of size $h_d {\times} w_d$ over the HR context crop $x_\mathit{c,HR}^T$ (see Fig~\ref{fig:architecture}\,b). The window is shifted with a stride of $\frac{h_d}{2} {\times} \frac{w_d}{2}$ to generate overlapping predictions with different contexts, which are averaged to increase robustness.

For inference, the full-scale HRDA semantic segmentation $\hat{y}_\mathit{F,HR}$ of the entire image $x_\mathit{HR}$ is necessary. As the context crop is usually smaller than the entire image, $\hat{y}_\mathit{F,HR}$ is generated using an overlapping sliding window over the entire image $x_\mathit{HR}$ with a size of $sh_c {\times} sw_c$ and a stride of $\frac{sh_c}{2} {\times} \frac{sw_c}{2}$. Within the sliding window, the HRDA prediction is generated in the same way as $\hat{y}_\mathit{c,F}^T$ for the pseudo-label.

\section{Experiments}
\label{sec:experiment}

\begin{table*}
\centering
\caption{Comparison of DAFormer and HRDA with state-of-the-art UDA methods. The performance is reported as IoU in \%. $\dagger$ indicates the use of additional daytime/clear-weather geographically-aligned reference images and $\ddagger$ denotes results reproduced with a MiT-B5 backbone.}
\label{tab:sota}
\setlength{\tabcolsep}{1.2pt}
\footnotesize
\begin{tabular}{l|c|ccccccccccccccccccc|c}
\toprule
Method & & Road & S.walk & Build. & Wall & Fence & Pole & Tr.Light & Sign & Vege. & Terrain & Sky & Person & Rider & Car & Truck & Bus & Train & M.bike & Bike & mIoU\\
\midrule
\multicolumn{21}{c}{\textbf{Synthetic-to-Real: GTA$\to$Cityscapes (Val.)}} \\
\midrule

AdaptSeg~\cite{tsai2018learning} & \parbox[t]{2mm}{\multirow{6}{*}{\rotatebox[origin=c]{90}{ResNet-101}}} & 86.5 & 25.9 & 79.8 & 22.1 & 20.0 & 23.6 & 33.1 & 21.8 & 81.8 & 25.9 & 75.9 & 57.3 & 26.2 & 76.3 & 29.8 & 32.1 & 7.2 & 29.5 & 32.5 & 41.4\\
ADVENT~\cite{vu2019advent} &  & 89.4 & 33.1 & 81.0 & 26.6 & 26.8 & 27.2 & 33.5 & 24.7 & 83.9 & 36.7 & 78.8 & 58.7 & 30.5 & 84.8 & 38.5 & 44.5 & 1.7 & 31.6 & 32.4 & 45.5\\
DACS~\cite{tranheden2021dacs} &  & 89.9 & 39.7 & 87.9 & 30.7 & 39.5 & 38.5 & 46.4 & 52.8 & 88.0 & 44.0 & 88.8 & 67.2 & 35.8 & 84.5 & 45.7 & 50.2 & 0.0 & 27.3 & 34.0 & 52.1\\
CorDA~\cite{wang2021domain} &  & 94.7 & 63.1 & 87.6 & 30.7 & 40.6 & 40.2 & 47.8 & 51.6 & 87.6 & 47.0 & 89.7 & 66.7 & 35.9 & 90.2 & 48.9 & 57.5 & 0.0 & 39.8 & 56.0 & 56.6\\
ProDA~\cite{zhang2021prototypical} &  & 87.8 & 56.0 & 79.7 & 46.3 & 44.8 & 45.6 & 53.5 & 53.5 & 88.6 & 45.2 & 82.1 & 70.7 & 39.2 & 88.8 & 45.5 & 59.4 & 1.0 & 48.9 & 56.4 & 57.5\\
DecoupleNet~\cite{lai2022decouplenet} &  & 87.6 & 49.3 & 87.2 & 42.5 & 41.6 & 46.6 & \underline{57.4} & 44.0 & 89.0 & 43.9 & 90.6 & \underline{73.0} & 43.8 & 88.1 & 32.9 & 53.7 & 44.3 & 49.8 & 57.2 & 59.1\\
\midrule
AdaptSeg$^\ddagger$~\cite{tsai2018learning} & \parbox[t]{2mm}{\multirow{4}{*}{\rotatebox[origin=c]{90}{MiT-B5}}} & 85.2 & 20.4 & 85.5 & 38.2 & 30.9 & 34.5 & 43.0 & 26.2 & 87.4 & 40.3 & 86.4 & 63.6 & 23.7 & 88.6 & 48.5 & 50.6 & 5.8 & 33.1 & 16.2 & 47.8\\
DACS$^\ddagger$~\cite{tranheden2021dacs} &  & 88.9 & 50.0 & 88.4 & 46.4 & 43.9 & 43.1 & 53.4 & 54.8 & \underline{89.9} & \textbf{51.2} & \underline{92.8} & 64.2 & 9.4 & 91.4 & \underline{77.3} & 63.3 & 0.0 & 47.4 & 49.8 & 58.2\\
DAFormer~(Ours) &  & \underline{95.7} & \underline{70.2} & \underline{89.4} & \underline{53.5} & \underline{48.1} & \underline{49.6} & 55.8 & \underline{59.4} & \underline{89.9} & 47.9 & 92.5 & 72.2 & \underline{44.7} & \underline{92.3} & 74.5 & \underline{78.2} & \underline{65.1} & \underline{55.9} & \underline{61.8} & \underline{68.3}\\
HRDA~(Ours) &  & \textbf{96.4} & \textbf{74.4} & \textbf{91.0} & \textbf{61.6} & \textbf{51.5} & \textbf{57.1} & \textbf{63.9} & \textbf{69.3} & \textbf{91.3} & \underline{48.4} & \textbf{94.2} & \textbf{79.0} & \textbf{52.9} & \textbf{93.9} & \textbf{84.1} & \textbf{85.7} & \textbf{75.9} & \textbf{63.9} & \textbf{67.5} & \textbf{73.8}\\

\midrule
\multicolumn{21}{c}{\textbf{Synthetic-to-Real: Synthia$\to$Cityscapes (Val.)}} \\
\midrule

ADVENT~\cite{vu2019advent} & \parbox[t]{2mm}{\multirow{5}{*}{\rotatebox[origin=c]{90}{ResNet}}} & 85.6 & 42.2 & 79.7 & 8.7 & 0.4 & 25.9 & 5.4 & 8.1 & 80.4 & -- & 84.1 & 57.9 & 23.8 & 73.3 & -- & 36.4 & -- & 14.2 & 33.0 & 41.2\\
DACS~\cite{tranheden2021dacs} &  & 80.6 & 25.1 & 81.9 & 21.5 & 2.9 & 37.2 & 22.7 & 24.0 & 83.7 & -- & \underline{90.8} & 67.6 & 38.3 & 82.9 & -- & 38.9 & -- & 28.5 & 47.6 & 48.3\\
CorDA~\cite{wang2021domain} &  & \textbf{93.3} & \textbf{61.6} & 85.3 & 19.6 & 5.1 & 37.8 & 36.6 & 42.8 & 84.9 & -- & 90.4 & 69.7 & 41.8 & 85.6 & -- & 38.4 & -- & 32.6 & 53.9 & 55.0\\
ProDA~\cite{zhang2021prototypical} &  & \underline{87.8} & 45.7 & 84.6 & 37.1 & 0.6 & 44.0 & 54.6 & 37.0 & \textbf{88.1} & -- & 84.4 & \underline{74.2} & 24.3 & 88.2 & -- & 51.1 & -- & 40.5 & 45.6 & 55.5\\
DecoupleNet~\cite{lai2022decouplenet} &  & 77.8 & \underline{48.6} & 75.6 & 32.0 & 1.9 & 44.4 & 52.9 & 38.5 & \underline{87.8} & -- & 88.1 & 71.1 & 34.3 & \underline{88.7} & -- & \underline{58.8} & -- & 50.2 & 61.4 & 57.0\\
\midrule
DACS$^\ddagger$~\cite{tranheden2021dacs} & \parbox[t]{2mm}{\multirow{3}{*}{\rotatebox[origin=c]{90}{MiT}}} & 58.0 & 46.0 & 84.8 & 37.7 & \underline{5.2} & 38.6 & 20.9 & 47.3 & 85.9 & -- & 81.6 & 73.0 & 43.9 & 86.9 & -- & 55.6 & -- & 51.1 & 18.6 & 52.2\\
DAFormer~(Ours) &  & 84.5 & 40.7 & \underline{88.4} & \underline{41.5} & \textbf{6.5} & \underline{50.0} & \underline{55.0} & \underline{54.6} & 86.0 & -- & 89.8 & 73.2 & \underline{48.2} & 87.2 & -- & 53.2 & -- & \underline{53.9} & \underline{61.7} & \underline{60.9}\\
HRDA~(Ours) &  & 85.2 & 47.7 & \textbf{88.8} & \textbf{49.5} & 4.8 & \textbf{57.2} & \textbf{65.7} & \textbf{60.9} & 85.3 & -- & \textbf{92.9} & \textbf{79.4} & \textbf{52.8} & \textbf{89.0} & -- & \textbf{64.7} & -- & \textbf{63.9} & \textbf{64.9} & \textbf{65.8}\\

\midrule
\multicolumn{21}{c}{\textbf{Day-to-Nighttime: Cityscapes$\to$DarkZurich (Test)}} \\
\midrule

ADVENT~\cite{vu2019advent} & \parbox[t]{2mm}{\multirow{4}{*}{\rotatebox[origin=c]{90}{ResNet}}} & 85.8 & 37.9 & 55.5 & 27.7 & 14.5 & 23.1 & 14.0 & 21.1 & 32.1 & 8.7 & 2.0 & 39.9 & 16.6 & 64.0 & 13.8 & 0.0 & 58.8 & 28.5 & 20.7 & 29.7\\
GCMA$^\dagger$~\cite{sakaridis2019guided} &  & 81.7 & 46.9 & 58.8 & 22.0 & \underline{20.0} & 41.2 & 40.5 & 41.6 & \underline{64.8} & \underline{31.0} & 32.1 & \underline{53.5} & 47.5 & 75.5 & 39.2 & 0.0 & 49.6 & 30.7 & 21.0 & 42.0\\
MGCDA$^\dagger$~\cite{sakaridis2020map} &  & 80.3 & 49.3 & 66.2 & 7.8 & 11.0 & 41.4 & 38.9 & 39.0 & 64.1 & 18.0 & 55.8 & 52.1 & \underline{53.5} & 74.7 & \underline{66.0} & 0.0 & 37.5 & 29.1 & 22.7 & 42.5\\
DANNet$^\dagger$~\cite{wu2021dannet} &  & 90.0 & 54.0 & \textbf{74.8} & \textbf{41.0} & \textbf{21.1} & 25.0 & 26.8 & 30.2 & \textbf{72.0} & 26.2 & \textbf{84.0} & 47.0 & 33.9 & 68.2 & 19.0 & 0.3 & 66.4 & 38.3 & 23.6 & 44.3\\
\midrule
DAFormer~(Ours) & \parbox[t]{2mm}{\multirow{2}{*}{\rotatebox[origin=c]{90}{MiT}}} & \textbf{93.5} & \textbf{65.5} & \underline{73.3} & 39.4 & 19.2 & \underline{53.3} & \underline{44.1} & \textbf{44.0} & 59.5 & \textbf{34.5} & 66.6 & 53.4 & 52.7 & \underline{82.1} & 52.7 & \underline{9.5} & \underline{89.3} & \underline{50.5} & \underline{38.5} & \underline{53.8}\\
HRDA~(Ours) &  & \underline{90.4} & \underline{56.3} & 72.0 & \underline{39.5} & 19.5 & \textbf{57.8} & \textbf{52.7} & \underline{43.1} & 59.3 & 29.1 & \underline{70.5} & \textbf{60.0} & \textbf{58.6} & \textbf{84.0} & \textbf{75.5} & \textbf{11.2} & \textbf{90.5} & \textbf{51.6} & \textbf{40.9} & \textbf{55.9}\\

\midrule
\multicolumn{21}{c}{\textbf{Clear-to-Adverse-Weather: Cityscapes$\to$ACDC (Test)}}  \\
\midrule

ADVENT~\cite{vu2019advent} & \parbox[t]{2mm}{\multirow{4}{*}{\rotatebox[origin=c]{90}{ResNet}}} & 72.9 & 14.3 & 40.5 & 16.6 & 21.2 & 9.3 & 17.4 & 21.2 & 63.8 & 23.8 & 18.3 & 32.6 & 19.5 & 69.5 & 36.2 & 34.5 & 46.2 & 26.9 & 36.1 & 32.7\\
GCMA$^\dagger$~\cite{sakaridis2019guided} &  & 79.7 & 48.7 & 71.5 & 21.6 & 29.9 & 42.5 & \underline{56.7} & \underline{57.7} & \textbf{75.8} & 39.5 & \textbf{87.2} & 57.4 & 29.7 & 80.6 & 44.9 & 46.2 & 62.0 & 37.2 & 46.5 & 53.4\\
MGCDA$^\dagger$~\cite{sakaridis2020map} &  & 73.4 & 28.7 & 69.9 & 19.3 & 26.3 & 36.8 & 53.0 & 53.3 & \underline{75.4} & 32.0 & 84.6 & 51.0 & 26.1 & 77.6 & 43.2 & 45.9 & 53.9 & 32.7 & 41.5 & 48.7\\
DANNet$^\dagger$~\cite{wu2021dannet} &  & \underline{84.3} & \underline{54.2} & 77.6 & 38.0 & 30.0 & 18.9 & 41.6 & 35.2 & 71.3 & 39.4 & \underline{86.6} & 48.7 & 29.2 & 76.2 & 41.6 & 43.0 & 58.6 & 32.6 & 43.9 & 50.0\\
\midrule
DAFormer~(Ours) & \parbox[t]{2mm}{\multirow{2}{*}{\rotatebox[origin=c]{90}{MiT}}} & 58.4 & 51.3 & \underline{84.0} & \underline{42.7} & \underline{35.1} & \underline{50.7} & 30.0 & 57.0 & 74.8 & \underline{52.8} & 51.3 & \underline{58.3} & \underline{32.6} & \underline{82.7} & \underline{58.3} & \underline{54.9} & \underline{82.4} & \underline{44.1} & \underline{50.7} & \underline{55.4}\\
HRDA~(Ours) &  & \textbf{88.3} & \textbf{57.9} & \textbf{88.1} & \textbf{55.2} & \textbf{36.7} & \textbf{56.3} & \textbf{62.9} & \textbf{65.3} & 74.2 & \textbf{57.7} & 85.9 & \textbf{68.8} & \textbf{45.7} & \textbf{88.5} & \textbf{76.4} & \textbf{82.4} & \textbf{87.7} & \textbf{52.7} & \textbf{60.4} & \textbf{68.0}\\

\bottomrule
\end{tabular}
\end{table*}

\begin{figure*}[tb]
\centering
\begin{minipage}{0.8\linewidth}
{\footnotesize
\begin{tabularx}{\linewidth}{*{5}{Y}}
Image & ProDA~\cite{zhang2021prototypical} & DAFormer (Ours) & HRDA (Ours) & Ground Truth \\
\end{tabularx}
} %
\includegraphics[width=\linewidth]{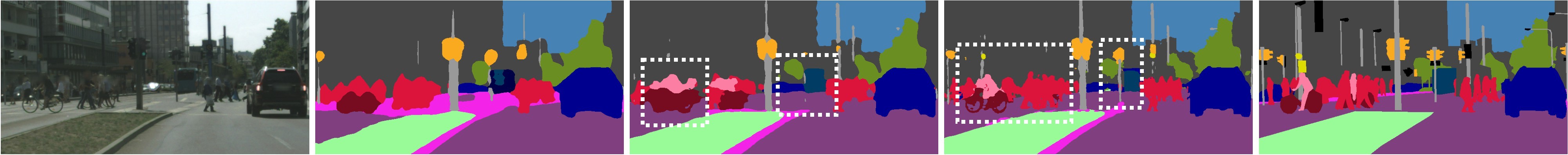}
\includegraphics[width=\linewidth]{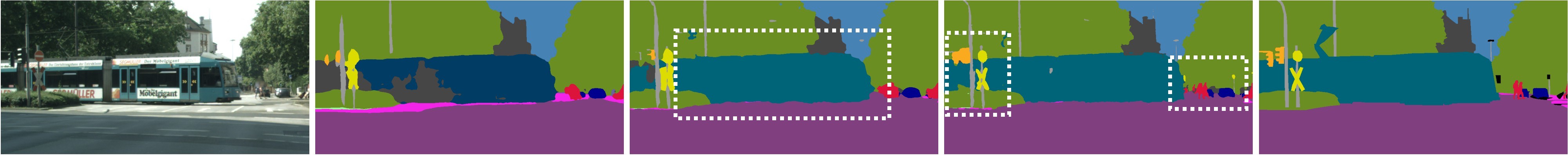}
{\footnotesize
\begin{tabularx}{\linewidth}{*{5}{Y}}
Image & DANNet~\cite{wu2021dannet} & DAFormer (Ours) & HRDA (Ours) & Ground Truth \\
\end{tabularx}
} %
\includegraphics[width=\linewidth]{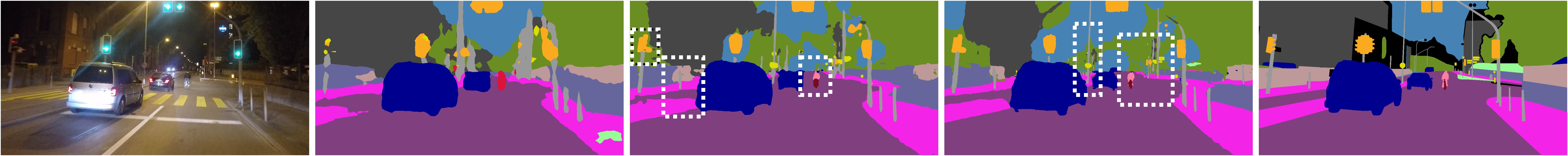}
\end{minipage}
\resizebox{0.805\linewidth}{!}{%
\scriptsize%
\setlength\tabcolsep{0.2pt}%
{%
\newcolumntype{P}[1]{>{\centering\arraybackslash}p{#1}}
\begin{tabular}{@{}*{20}{P{0.09\columnwidth}}@{}}
     {\cellcolor[rgb]{0.5,0.25,0.5}}\textcolor{white}{road} 
     &{\cellcolor[rgb]{0.957,0.137,0.91}}sidew. 
     &{\cellcolor[rgb]{0.275,0.275,0.275}}\textcolor{white}{build.} 
     &{\cellcolor[rgb]{0.4,0.4,0.612}}\textcolor{white}{wall} 
     &{\cellcolor[rgb]{0.745,0.6,0.6}}fence 
     &{\cellcolor[rgb]{0.6,0.6,0.6}}pole 
     &{\cellcolor[rgb]{0.98,0.667,0.118}}tr. light
     &{\cellcolor[rgb]{0.863,0.863,0}}tr. sign 
     &{\cellcolor[rgb]{0.42,0.557,0.137}}veget. 
     &{\cellcolor[rgb]{0.596,0.984,0.596}}terrain 
     &{\cellcolor[rgb]{0.275,0.510,0.706}}sky
     &{\cellcolor[rgb]{0.863,0.078,0.235}}\textcolor{white}{person} 
     &{\cellcolor[rgb]{0.988,0.494,0.635}}\textcolor{black}{rider} 
     &{\cellcolor[rgb]{0,0,0.557}}\textcolor{white}{car} 
     &{\cellcolor[rgb]{0,0,0.275}}\textcolor{white}{truck} 
     &{\cellcolor[rgb]{0,0.235,0.392}}\textcolor{white}{bus}
     &{\cellcolor[rgb]{0,0.392,0.471}}\textcolor{white}{train} 
     &{\cellcolor[rgb]{0,0,0.902}}\textcolor{white}{m.bike} 
     & {\cellcolor[rgb]{0.467,0.043,0.125}}\textcolor{white}{bike}
     &{\cellcolor[rgb]{0,0,0}}\textcolor{white}{n/a.}
\end{tabular}
}%
}%
\caption{Qualitative comparison of DAFormer and HRDA with previous methods on GTA$\rightarrow$Cityscapes and Cityscapes$\to$DarkZurich UDA. DAFormer better distinguishes difficult classes such as pedestrian, rider, bus, and train. HRDA further improves the segmentation of small instances.}
\label{fig:visual_examples}
\end{figure*}

\begin{figure*}[tb]
\centering
\begin{minipage}{0.8\linewidth}
{\footnotesize
\begin{tabularx}{\linewidth}{*{5}{Y}}
Image & SHADE~\cite{zhao2022style} & DAFormer (Ours) & HRDA (Ours) & Ground Truth \\
\end{tabularx}
} %
\includegraphics[width=\linewidth]{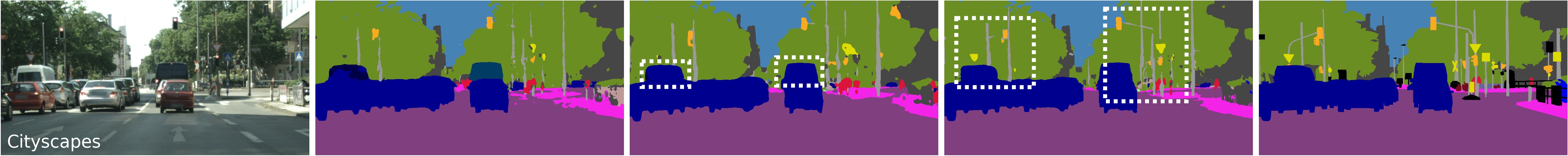}
\includegraphics[width=\linewidth]{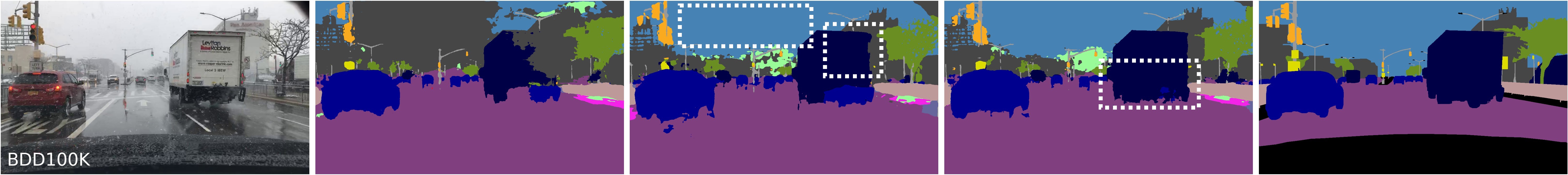}
\includegraphics[width=\linewidth]{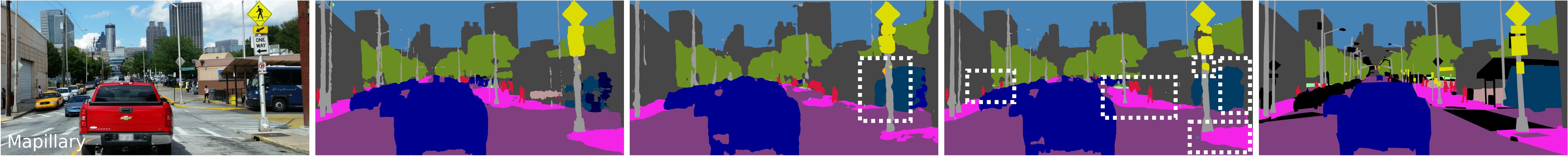}
\end{minipage}
\caption{Qualitative comparison of SHADE, DAFormer, and HRDA for DG trained on GTA and evaluated on Cityscapes, BDD100K, and Mapillary.}
\label{fig:visual_examples_dg}
\end{figure*}

\subsection{Implementation Details}

\noindent{\textbf{Datasets}}
As synthetic datasets, we utilize GTA~\cite{richter2016playing} with 24,966 images of 1914$\times$1052 pixels and Synthia~\cite{ros2016synthia} with 9,400 images of 1280$\times$760 pixels. As real-world datasets, we use Cityscapes~\cite{cordts2016cityscapes} with 2,975 training and 500 validation images of 2048${\times}$1024 pixels for clear weather, DarkZurich~\cite{sakaridis2020map} with 2,416 training, 50 validation, and 151 test images of 1920${\times}$1080 pixels for nighttime, and ACDC~\cite{sarkadis2021acdc} with 1,600 training, 406 validation, and 2,000 test images of 1920${\times}$1080 pixels for adverse weather (fog, night, rain, and snow). For DG, we additionally evaluate the model on the US-focused BDD100K~\cite{yu2020bdd100k} and the world-wide Mapillary Vistas~\cite{neuhold2017mapillary} with 1000 and 2000 validation images, respectively.
Both contain different illumination (day/night) and weather conditions (mostly sunny/cloudy but also rainy/snowy/foggy).
Previous works~\cite{tsai2018learning,tranheden2021dacs,araslanov2021self} downsample Cityscapes to 1024${\times}$512 and GTA to 1280${\times}$720. Instead, HRDA maintains the full resolution for Cityscapes. To train with the same scale ratio of source and target images as previous works, HRDA resizes GTA to 2560${\times}$1440 and Synthia to 2560${\times}$1520 pixels.

\noindent{\textbf{Network Architecture}}
Our implementation is based on the mmsegmentation framework~\cite{mmseg2020}. For the DAFormer architecture, we use the MiT-B5 encoder~\cite{xie2021segformer}, which produces a feature pyramid with $C=[64, 128, 320, 512]$. All encoders are pretrained on ImageNet-1k. The DAFormer decoder uses $C_e=256$ and dilation rates of 1, 6, 12, and 18.
The scale attention decoder of HRDA uses the lightweight SegFormer MLP decoder~\cite{xie2021segformer} with an embedding dimension of 256.

\noindent{\textbf{Training}}
In accordance with~\cite{xie2021segformer}, we train DAFormer and HRDA with AdamW, a learning rate of $\eta_\mathit{base} {=} 6 {\times} 10^{-5}$ for the encoder and $ 6 {\times} 10^{-4}$ for the decoder, a weight decay of $0.01$, linear learning rate warmup with $t_\mathit{warm}{=}1.5$k, and linear decay afterwards. It is trained with a batch size of $2$ for $40$k iterations.
We use the same data augmentation parameters as DACS~\cite{tranheden2021dacs} and set $\alpha{=}0.999$ and $\tau{=}0.968$.
The RCS temperature is set $T{=}0.01$ to maximize the sampled pixels of the class with the least pixels.
For FD, $r{=}0.75$ and $\lambda_\mathit{FD} {=} 0.005$ to induce a similar gradient magnitude into the encoder as $\mathcal{L}_S$.
The context and detail crop are generated using $h_c{=}w_c{=}h_d{=}w_d{=}512$ with $s{=}2$ to balance the required resources for both crops in the default case. The detail loss weight is chosen empirically $\lambda_d{=}0.1$. 
For DG, we follow the parameters of SHADE~\cite{zhao2022style}.
The experiments are conducted on an RTX 2080 Ti GPU with 11 GB memory (for DAFormer) or a Titan RTX GPU with 24 GB (for HRDA). The mIoUs are calculated with the last checkpoint of the student model $f_\theta$ (no model selection) and are averaged over 3 random seeds.

\subsection{Comparison with State-of-the-Art UDA and DG}

\begin{table}
\centering
\caption{
Comparison of DG methods trained on GTA and evaluated on the validation set of Cityscapes (CS), BDD100K, Mapillary (Map.), ACDC, and DarkZurich (DZ). $\dagger$ indicates results with ResNet-101 obtained from \cite{zhao2022style} and $\ddagger$ denotes results reproduced with a MiT-B5 backbone.
}
\label{tab:sota_dg}
\setlength{\tabcolsep}{1.5pt}
\footnotesize
\begin{tabular}{l|c|cccc|ccc}
\toprule
DG Method & & CS & BDD & Map. & Avg$_3$ & ACDC & DZ & Avg$_5$\\
\midrule
PhotoDistort\,(Sec.\,\ref{sec:methods_dg}) & \parbox[t]{2mm}{\multirow{9}{*}{\rotatebox[origin=c]{90}{ResNet-101}}} & 34.34 & 32.43 & 31.41 & 32.73 & 24.20 & 5.49 & 25.58\\
IBN-Net$^\dagger$~\cite{pan2018two} &  & 37.37 & 34.21 & 36.81 & 36.13 & -- & -- & --\\
DRPC~\cite{yue2019domain} &  & 42.53 & 38.72 & 38.05 & 39.77 & -- & -- & --\\
ISW$^\dagger$~\cite{choi2021robustnet} &  & 37.20 & 33.36 & 35.57 & 35.38 & -- & -- & --\\
FSDR~\cite{huang2021fsdr} &  & 44.80 & 41.20 & 43.40 & 43.13 & 24.77 & 9.66 & 32.77\\
GTR~\cite{peng2021global} &  & 43.70 & 39.60 & 39.10 & 40.80 & -- & -- & --\\
SAN-SAW~\cite{peng2022semantic} &  & 45.33 & 41.18 & 40.77 & 42.43 & -- & -- & --\\
AdvStyle~\cite{zhongadversarial} &  & 44.51 & 39.27 & 43.48 & 42.42 & -- & -- & --\\
SHADE~\cite{zhao2022style} &  & 46.66 & 43.66 & 45.50 & 45.27 & 29.06 & 8.01 & 34.58\\
\midrule
PhotoDistort$^\ddagger$\,(Sec.\,\ref{sec:methods_dg}) & \parbox[t]{2mm}{\multirow{4}{*}{\rotatebox[origin=c]{90}{MiT-B5}}} & 45.57 & 43.02 & 47.51 & 45.36 & 35.15 & 15.55 & 37.36\\
SHADE$^\ddagger$~\cite{zhao2022style} &  & 48.52 & 45.17 & 51.69 & 48.46 & 36.33 & 16.21 & 39.58\\
DAFormer &  & \underline{52.65} & \underline{47.89} & \underline{54.66} & \underline{51.73} & \underline{38.25} & \underline{17.45} & \underline{42.18}\\
HRDA &  & \textbf{57.41} & \textbf{49.11} & \textbf{61.16} & \textbf{55.9} & \textbf{44.04} & \textbf{20.97} & \textbf{46.54}\\

\bottomrule
\end{tabular}
\end{table}

Tab.~\ref{tab:sota} shows that the proposed DAFormer and HRDA outperform previous UDA methods by a significant margin.
For synthetic-to-real UDA, they improve the performance from 57.5 to 73.8 mIoU on GTA$\rightarrow$Cityscapes and from 55.5 to 65.8 mIoU on Synthia$\rightarrow$Cityscapes. They also outperform concurrent methods such as DecoupleNet~\cite{lai2022decouplenet}.
For day-to-nighttime UDA on Cityscapes$\to$DarkZurich, DAFormer and HRDA enhance the performance from 44.3 to 55.9 mIoU. And for clear-to-adverse-weather UDA on Cityscapes$\to$ACDC, they increase the mIoU from 50.0 to 68.0. In contrast to most previous works in this setting, DAFormer and HRDA do not require paired clear-weather/adverse-weather images of the same locations for the adaptation process.
The class-wise IoUs in Tab.~\ref{tab:sota} show that DAFormer and HRDA learn even difficult classes well, which previous methods struggled with such as train, bus, and truck. HRDA further achieves significant improvements for classes with fine segmentation details such as pole, traffic light, traffic sign, person, rider, motorbike, and bike. But also the other classes benefit from the multi-resolution UDA. These observations are also reflected in the visual examples in Fig.~\ref{fig:visual_examples}.

Further, Tab.~\ref{tab:sota_dg} shows that DAFormer and HRDA outperform previous DG methods by +10.8/\allowbreak +5.5/\allowbreak +15.7/+15.0/11.3 when generalizing from GTA to Cityscapes/\allowbreak BDD100K/\allowbreak Mapillary/\allowbreak ACDC/\allowbreak DarkZurich, respectively. The smaller improvement on BDD100K is probably due to its lower resolution, limiting the potential of HRDA. Fig.~\ref{fig:visual_examples_dg} shows example predictions for common improvements.

Tab.~\ref{tab:sota} and \ref{tab:sota_dg} further provide results of previous methods reproduced with the Transformer MiT-B5 backbone instead of a CNN ResNet. It can be seen that MiT-B5 generally improves the mIoU across the tested DG and UDA methods. Still, the other contributions of DAFormer and HRDA (such as FD, RCS, and multi-scale/crop training) provide further significant gains as will be discussed in Sec. \ref{sec:component_ablation} - \ref{sec:exp_multi_resolution_uda}.

These results display that DAFormer and HRDA are widely applicable to learn domain-robust semantic segmentation and to bridge various domain shifts.
Beyond the presented benchmarks, DAFormer/HRDA were also extended to further use cases such as 
waste sorting~\cite{bashkirova2023visda},
panoptic segmentation~\cite{saha2023edaps},
remote sensing~\cite{xia2023openearthmap}, %, li2022unsupervised
and aerial power line segmentation~\cite{rao2022quadformer}, where they achieve strong performances.

\subsection{Component Ablation Study for DG and UDA}
\label{sec:component_ablation}

\begin{table}
\centering
\caption{
Ablation study of DAFormer and HRDA: Learning rate warmup (W), Transformer encoder (Transf.), ImageNet Thing-Class Feature Distance (FD), Rare Class Sampling (RCS), stabilized UDA self-training (Stab.), DAFormer decoder (DAF.), multi-resolution training (HRDA), and SHADE~\cite{zhao2022style}. The mIoU (\%) is reported for DG and UDA on GTA$\to$Cityscapes. Mean and SD are calculated over 3 random seeds.
}
\label{tab:highlevel_ablation}
\setlength{\tabcolsep}{2pt}
\footnotesize

\begin{tabular}{llllllllll}
\toprule
W & Transf. & FD& RCS & Stab. & DAF. & HRDA & SHADE & DG & UDA \\
\midrule
-- & --  & -- & --  & -- & --  & -- & --  & 25.2\spm{1.6} & 49.1\spm{2.0} \\
\cm & --  & -- & --  & -- & --  & -- & --  & 34.3\spm{2.2} & 54.2\spm{1.7} \\
\cm & \cm & --  & -- & --  & -- & --  & -- & 45.6\spm{0.6} & 58.2\spm{0.9} \\
\midrule
\g \cm & \g \cm & \g $\mathcal{C}_\mathit{all}$ & \g -- & \g -- & \g -- & \g -- & \g -- & \g 48.2\spm{1.3} & \g 58.8\spm{0.4} \\
\cm & \cm & \cm & --  & -- & --  & -- & --  & 49.6\spm{0.6} & 61.7\spm{2.6} \\
\g \cm & \g \cm & \g --  & \g $T{=}\infty$ & \g --  & \g -- & \g --  & \g -- & \g 48.7\spm{0.4}  & \g 62.0\spm{1.5} \\
\cm & \cm & --  & \cm & --  & -- & --  & -- & 50.0\spm{1.2} & 64.0\spm{2.4} \\
\cm & \cm & \cm & \cm & --  & -- & --  & -- & 50.7\spm{0.3} & 66.2\spm{1.0} \\
\cm & \cm & \cm & \cm & \cm & --  & -- & --  & --\texttt{"}-- & 67.0\spm{0.4} \\
\midrule
\cm & \cm & \cm & \cm & \cm & \cm & --  & -- & 51.6\spm{1.1} & 68.3\spm{0.5} \\
\cm & \cm & \cm & \cm & \cm & \cm & \cm & --  & 55.5\spm{0.5} & 73.8\spm{0.3} \\
\midrule
\g \cm & \g \cm & \g \cm & \g -- & \g -- & \g -- & \g --  & \g \cm & \g 48.5\spm{1.0} & \g --- \\
\cm & \cm & \cm & \cm & \cm & \cm & --  & \cm & 52.6\spm{1.1} & --- \\
\cm & \cm & \cm & \cm & \cm & \cm & \cm & \cm & 57.4\spm{0.4} & --- \\
\bottomrule
\end{tabular}

\end{table}

Tab.~\ref{tab:highlevel_ablation} shows the high-level ablation study of the proposed components of DAFormer and HRDA with their DG and UDA performances on GTA$\to$Cityscapes (mIoU in \%). As initial baseline, DG with photometric distortion for style augmentation (see Sec.~\ref{sec:methods_dg}) and UDA self-training (see Sec.~\ref{sec:methods_self_training}) are used.
Each of the proposed components results in successive significant improvements of the DG and UDA performance. In particular, learning rate warmup improves the DG/UDA performance by +9.1/+5.1 (see Sec.~\ref{sec:exp_lr_warmup} for more details). The Transformer encoder further improves the domain robustness resulting in a gain of +11.3/+4.0 (see Sec.~\ref{sec:exp_comparison_networks}). Regularizing the training with the ImageNet FD further increases the DG/UDA mIoU by +4.0/+3.5 (see Sec.~\ref{sec:exp_fd}) while RCS enhances it by +4.4/+5.8 (see Sec.~\ref{sec:exp_rcs}). When combining RCS and FD, we observe an additional improvement by +0.7/+2.2 showing that they complement each other. At this point, we notice that the UDA performance is limited by pseudo-label drifts originating from image rectification artifacts and the ego car. As these regions are not part of the street scene segmentation task, we argue that it is not meaningful to produce pseudo-labels for them. 
To stabilize the UDA self-training, we ignore the top 15 and bottom 120 pixels of the pseudo-label. As Transformers are more expressive, we further increase $\alpha$ to 0.999 to introduce a stronger regularization from the teacher. This mitigates the pseudo-label drifts and improves the UDA performance by +0.8 mIoU. Based on the stabilized UDA, we study the decoder architecture in Sec.~\ref{sec:exp_context_aware_fusion}. The proposed DAFormer decoder with context-aware feature fusion further improves the DG/UDA mIoU by +0.9/+1.3. The multi-resolution training with HRDA results in another major gain of +3.9/+5.5 (see Sec.~\ref{sec:exp_multi_resolution_uda}). 
Finally, we add the more advanced SHADE~\cite{zhao2022style} style diversification and consistency (see Sec.~\ref{sec:methods_dg}), which additionally increases the DG performance by +1.0 for DAFormer and +1.9 for HRDA.
In the following, we analyze the different components in more detail.

\subsection{Network Architecture Study}
\label{sec:exp_comparison_networks}

\begin{table}
\centering
\caption{
Comparison of different segmentation architectures wrt. their DG, UDA, and oracle performance.
Additionally, the relative UDA performance $\mathit{Rel.} = \frac{\mathit{UDA}}{\mathit{Oracle}}$ is provided. 
}
\label{tab:basic_architecture_comparison}
\footnotesize
\begin{tabular}{lllll}
\toprule
    Architecture &       DG &           UDA &         Oracle &   Rel. \\
\midrule
    DeepLabV2~\cite{chen2017deeplab} & 34.3\spm{2.2} & 54.2\spm{1.7} & 72.1\spm{0.5} & 75.2\% \\
        DA Net~\cite{fu2019dual} & 30.9\spm{2.1} & 53.7\spm{0.2} & 72.6\spm{0.2} & 74.0\% \\
        ISA Net~\cite{huang2019interlaced} & 32.3\spm{2.1} & 53.3\spm{0.4} & 72.0\spm{0.5} & 74.0\% \\
    DeepLabV3+~\cite{chen2018encoder} & 31.0\spm{1.4} & 53.7\spm{1.0} & 75.6\spm{0.9} & 71.0\% \\
    SegFormer~\cite{xie2021segformer} & \textbf{45.6}\spm{0.6} & \textbf{58.2}\spm{0.9} & \textbf{76.4}\spm{0.2} & \textbf{76.2\%} \\
\bottomrule
\end{tabular}
\end{table}

\begin{table}
\centering
\caption{Ablation of the SegFormer encoder and decoder.}
\label{tab:encoder_decoder}
\footnotesize
\setlength{\tabcolsep}{5pt}
\begin{tabular}{llllll}
\toprule
    Encoder &             Decoder &           DG & UDA &         Oracle &   Rel. \\
\midrule
    MiT-B5~\cite{xie2021segformer} &           SegF.~\cite{xie2021segformer} & 45.6\spm{0.6} & 58.2\spm{0.9} & 76.4\spm{0.2} & 76.2\% \\
    MiT-B5~\cite{xie2021segformer} &           DLv3+~\cite{chen2018encoder} & 45.3\spm{1.4} & 56.8\spm{1.8} & 75.5\spm{0.5} & 75.2\% \\
    R101~\cite{he2016deep} &           SegF.~\cite{xie2021segformer} & 31.9\spm{2.2} & 50.9\spm{1.1} & 71.3\spm{1.3} & 71.4\% \\
    R101~\cite{he2016deep} &           DLv3+~\cite{chen2018encoder} & 31.0\spm{1.4} & 53.7\spm{1.0} & 75.6\spm{0.9} & 71.0\% \\
\bottomrule
\end{tabular}
\end{table}

As a foundation for the DAFormer architecture, we compare several semantic segmentation architectures with respect to their DG, UDA, and oracle performance.
The oracle is trained on the labeled target domain, which gives a measure of the capacity of the architecture for supervised learning. 
To compare how well a network is suited for UDA, we further provide the relative performance $\mathit{Rel.} {=} \frac{\mathit{UDA}}{\mathit{Oracle}}$. 
Tab.~\ref{tab:basic_architecture_comparison} shows that a higher oracle performance does not necessarily increase the DG or UDA performance as can be seen for DeepLabV3+. Generally, the studied more recent CNN architectures, do not provide a DG or UDA performance gain over DeepLabV2. However, we identified the Transformer-based SegFormer as a powerful architecture for DG and UDA.

To analyze why SegFormer works well for DG and UDA, we swap its encoder and decoder with ResNet101 and DeepLabV3+.
Tab.~\ref{tab:encoder_decoder} shows that the lightweight MLP decoder of SegFormer has a slightly higher relative UDA performance (Rel.) than the heavier DLv3+ decoder (76.2\% vs. 75.2\%). However, the crucial contribution comes from the Transformer MiT encoder. Replacing it with the ResNet101 encoder leads to a significant performance drop of the DG and UDA performance. Even though the oracle performance also slightly decreases,
the drop for DG and UDA is over-proportional so that Rel. decreases from 76.2\% to 71.4\%.

\begin{table}
\centering
\caption{Influence of the encoder on the DG and UDA performance.}
\label{tab:encoder_size}
\footnotesize
\begin{tabular}{llllll}
\toprule
Encoder & Decoder &       DG &           UDA &         Oracle &   Rel. \\
\midrule
    R50~\cite{he2016deep} &    DLv2~\cite{chen2017deeplab} &     29.3 & 52.1 &   70.8 & 73.6\% \\
    R101~\cite{he2016deep} &    DLv2~\cite{chen2017deeplab} &     36.9 & 53.3 &   72.5 & 73.5\% \\
    S50~\cite{zhang2020resnest} &    DLv2~\cite{chen2017deeplab} &     27.9 & 48.0 &   67.7 & 70.9\% \\
    S101~\cite{zhang2020resnest} &    DLv2~\cite{chen2017deeplab} &     35.5 & 53.5 &   72.2 & 74.1\% \\
    S200~\cite{zhang2020resnest} &    DLv2~\cite{chen2017deeplab} &     35.9 & 56.9 &   73.5 & 77.4\% \\
    MiT-B3~\cite{xie2021segformer} &   SegF.~\cite{xie2021segformer} &     42.2 & 50.8 &   76.5 & 66.4\% \\
    MiT-B4~\cite{xie2021segformer} &   SegF.~\cite{xie2021segformer} &     44.7 & 57.5 &   77.1 & 74.6\% \\
    MiT-B5~\cite{xie2021segformer} &   SegF.~\cite{xie2021segformer} &     46.2 & 58.8 &   76.2 & 77.2\% \\
\bottomrule
\end{tabular}
\end{table}

\begin{figure}
\centering
\input{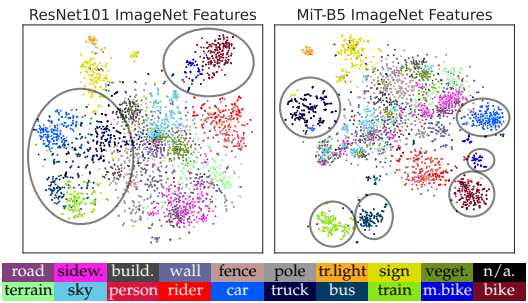}
\vspace{-0.3cm}
\resizebox{0.81\linewidth}{!}{%
\scriptsize
\setlength\tabcolsep{1pt}
{
\newcolumntype{P}[1]{>{\centering\arraybackslash}p{#1}}
\begin{tabular}{@{}*{10}{P{0.081\columnwidth}}@{}}
     {\cellcolor[rgb]{0.5,0.25,0.5}}\textcolor{white}{road} &{\cellcolor[rgb]{0.957,0.137,0.91}}sidew. &{\cellcolor[rgb]{0.275,0.275,0.275}}\textcolor{white}{build.} &{\cellcolor[rgb]{0.4,0.4,0.612}}\textcolor{white}{wall} &{\cellcolor[rgb]{0.745,0.6,0.6}}fence &{\cellcolor[rgb]{0.6,0.6,0.6}}pole &{\cellcolor[rgb]{0.98,0.667,0.118}}tr.light&{\cellcolor[rgb]{0.863,0.863,0}}sign &{\cellcolor[rgb]{0.42,0.557,0.137}}veget. & {\cellcolor[rgb]{0,0,0}}\textcolor{white}{n/a.}\\
     
     {\cellcolor[rgb]{0.596,0.984,0.596}}terrain &{\cellcolor[rgb]{0.392,0.784,0.902}}sky &{\cellcolor[rgb]{0.863,0.078,0.235}}\textcolor{white}{person} &{\cellcolor[rgb]{1,0,0}}\textcolor{white}{rider} &{\cellcolor[rgb]{0,0.353,1}}\textcolor{white}{car} &{\cellcolor[rgb]{0,0,0.275}}\textcolor{white}{truck} &{\cellcolor[rgb]{0,0.235,0.392}}\textcolor{white}{bus}&{\cellcolor[rgb]{0.549,0.902,0.078}}\textcolor{black}{train} &{\cellcolor[rgb]{0,0,0.902}}\textcolor{white}{m.bike} & {\cellcolor[rgb]{0.467,0.043,0.125}}\textcolor{white}{bike}\\
\end{tabular}
}
}
\vspace{0.2cm}
\caption{T-SNE embedding of the bottleneck features after ImageNet pretraining for ResNet101~\cite{he2016deep} and MiT-B5~\cite{xie2021segformer} on the Cityscapes validation set, showing a better vehicle separability for MiT.}
\label{fig:tsne_imagenet}
\end{figure}

Therefore, we further investigate the influence of the encoder architecture on the DG and UDA performance. Tab.~\ref{tab:encoder_size} shows that deeper models achieve a better DG, UDA, and Rel. performance demonstrating their improved ability to generalize/adapt to a new domain.
Overall, the best DG and UDA mIoUs are achieved by the Transformer MiT-B5 encoder. 
To gain insights on the improved generalization, Fig.~\ref{fig:tsne_imagenet} visualizes the features of the ImageNet pre-trained model inferred on the target domain (Cityscapes). Even though ResNet structures stuff-classes slightly better, MiT shines at separating semantically similar classes (e.g. all vehicle classes), which are usually particularly difficult to generalize/adapt. A possible explanation might be the texture-bias of CNNs and the shape-bias of Transformers~\cite{bhojanapalli2021understanding}.

\subsection{Learning Rate Warmup (W)}
\label{sec:exp_lr_warmup}

\begin{table}
\centering
\caption{Influence of learning rate warmup on the DG and UDA performance.}
\label{tab:warmup}
\setlength{\tabcolsep}{3pt}
\footnotesize
\begin{tabular}{llllll}
\toprule
Architecture & LR Warmup &           DG & UDA &         Oracle &   Rel. \\
\midrule
    DLv2~\cite{chen2017deeplab} &      -- & 25.2\spm{1.6} & 49.1\spm{2.0} & 67.4\spm{1.7} & 72.8\% \\
    DLv2~\cite{chen2017deeplab} &     \cm & 34.3\spm{2.2} & 54.2\spm{1.7} & 72.1\spm{0.5} & 75.2\% \\
    SegF.~\cite{xie2021segformer} &    -- & 35.5\spm{3.9} & 51.8\spm{0.8} & 72.9\spm{1.6} & 71.1\% \\
    SegF.~\cite{xie2021segformer} &   \cm & 45.6\spm{0.6} & 58.2\spm{0.9} & 76.4\spm{0.2} & 76.2\% \\
\bottomrule
\end{tabular}
\end{table}

\begin{figure}
\centering
\includegraphics[width=\linewidth]{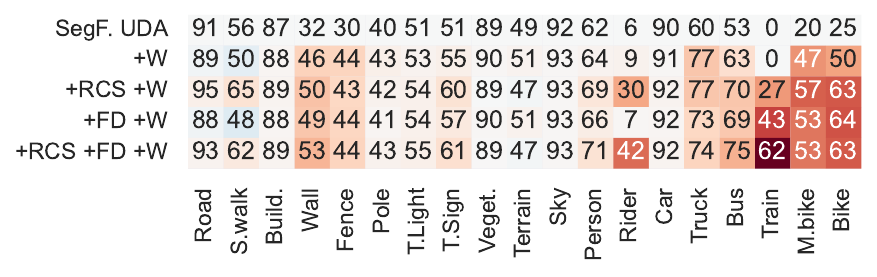}
\vspace{-0.8cm}
\caption{Comparison of the class-wise IoU of Warmup (W), RCS and FD for UDA. The color visualizes the IoU difference to the baseline.}
\label{fig:ciou_heatmap}
\end{figure}

Tab.~\ref{tab:warmup} shows that learning rate warmup significantly improves DG, UDA, and oracle mIoU. 
DG\&UDA benefit even more than supervised learning (see Rel.), showing its particular importance for DG\&UDA by stabilizing the beginning of the training, which improves difficult classes (see Fig~\ref{fig:ciou_heatmap}).

\begin{figure}
\centering
\includegraphics[width=\linewidth]{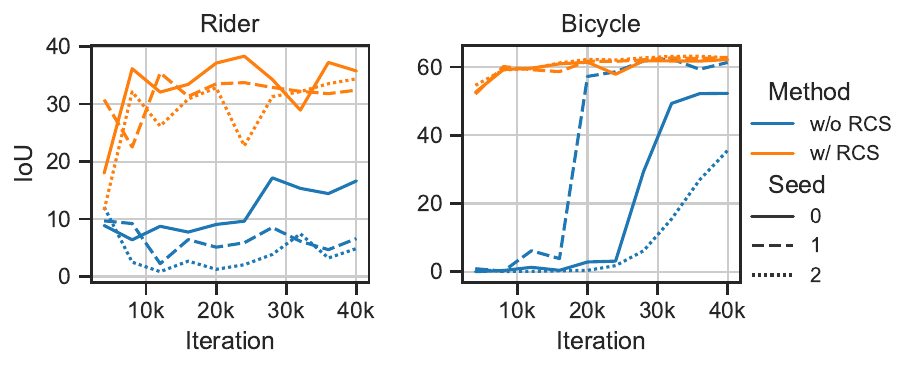}
\vspace{-0.8cm}
\caption{SegFormer UDA performance for the rare classes rider and bicycle without and with Rare Class Sampling (RCS).}
\label{fig:RCS_iou}
\end{figure}
\begin{figure}
\centering
\includegraphics[width=\linewidth]{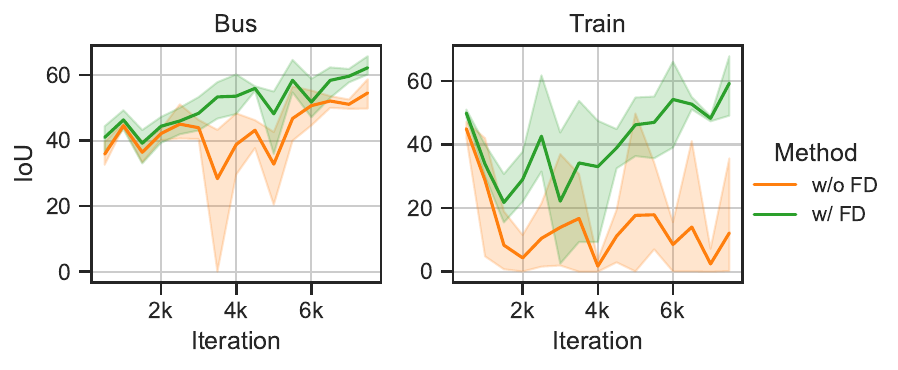}
\vspace{-0.8cm}
\caption{SegFormer UDA performance in the beginning of the training with and without ImageNet Feature Distance (FD).}
\label{fig:fd_iou}
\end{figure}

\subsection{Rare Class Sampling (RCS)}
\label{sec:exp_rcs}

The blue IoU curves in Fig.~\ref{fig:RCS_iou} indicate that the IoU of some classes depends on the random seed for data sampling. These classes are underrepresented in the source dataset. Interestingly, the IoU for the class bicycle starts increasing at different iterations for different seeds. We hypothesize that this is caused by the sampling order, in particular when the relevant rare classes are sampled. Further, the later the IoU starts improving, the worse is the final IoU of this class, probably due to the confirmation bias of self-training that was accumulated over earlier iterations.
Therefore, for UDA, it is especially important to learn rare classes early.
To address this issue, the proposed RCS increases the sampling probability of rare classes.
Fig.~\ref{fig:RCS_iou} (orange) shows that RCS results in an earlier increase of the IoU of rider/bicycle and a higher final IoU independent of the data sampling random seed. This confirms the hypothesis that an (early) sampling of rare classes is important for learning these classes properly. 
RCS improves the DG/UDA performance by +4.4/+5.8 mIoU (see Tab.~\ref{tab:highlevel_ablation}).
The highest IoU increase is observed for the rare classes rider, train, motorcycle, and bicycle (see Fig.~\ref{fig:ciou_heatmap}).
RCS also outperforms its special case `class-balanced sampling' with $T {=} \infty$ by +1.3/+2.0 mIoU for DG/UDA (see Tab.~\ref{tab:highlevel_ablation}) as class-balanced sampling does not consider the co-occurrence of multiple classes in semantic segmentation.

\subsection{Thing-Class ImageNet Feature Distance (FD)}
\label{sec:exp_fd}

Even though thing-classes are fairly well separated in ImageNet features (see Fig.~\ref{fig:tsne_imagenet} right), some of them such as bus and train are mixed together after DG or UDA training.
When investigating the IoU during the early training (see Fig.~\ref{fig:fd_iou} orange), we observe an early performance drop for the class train. We assume that the powerful MiT encoder overfits to the source domain. When regularizing the training with the proposed FD, the performance drop is avoided (see Fig.~\ref{fig:fd_iou} green). 
Also other difficult classes such as bus, motorcycle, and bicycle benefit from the regularization (see Fig.~\ref{fig:ciou_heatmap}). Tab.~\ref{tab:highlevel_ablation} shows that FD improves the DG/UDA mIoU by +4.0/+3.5. Note that applying FD only to thing-classes, which the ImageNet features were trained on, is important for a good performance (+1.4/+2.9 mIoU).

\begin{table}
\centering
\caption{Comparison of decoder architectures with MiT encoder and UDA improvements (DSC: depthwise separable convolution).}
\label{tab:decoder_fusion}
\footnotesize
\setlength{\tabcolsep}{3.5pt}
\begin{tabular}{lcrccc}
\toprule
            Decoder & $C_e$ & \#Params &          UDA &         Oracle &   Rel. \\
\midrule
        SegF.~\cite{xie2021segformer} & 768 &  3.2M & 67.0\spm{0.4} & 76.8\spm{0.3} & 87.2\% \\
        SegF.~\cite{xie2021segformer} & 256 &  0.5M & 67.1\spm{1.1} & 76.5\spm{0.4} & 87.7\% \\
        UperNet~\cite{xiao2018unified} & 512 & 29.6M & 67.4\spm{1.1} & \textbf{78.0}\spm{0.2} & 86.4\% \\
        UperNet~\cite{xiao2018unified} & 256 &  8.3M & 66.7\spm{1.2} & 77.4\spm{0.3} & 86.2\% \\
ISA~\cite{huang2019interlaced} Fusion & 256 &  1.1M & 66.3\spm{0.9} & 76.3\spm{0.4} & 86.9\% \\
Context only at $F_4$ & 256 & 3.2M & 67.0\spm{0.6} & 76.6\spm{0.2} & 87.5\% \\
                        DAFormer w/o DSC & 256 & 10.0M & 67.0\spm{1.5} & 76.7\spm{0.6} & 87.4\% \\
                DAFormer & 256 &  3.7M & \textbf{68.3}\spm{0.5} & 77.6\spm{0.2} & \textbf{88.0\%} \\
\bottomrule
\end{tabular}
\end{table}

\subsection{DAFormer Decoder}
\label{sec:exp_context_aware_fusion}

After regularizing and stabilizing the DG and UDA training, we come back to the network architecture and investigate the DAFormer decoder with context-aware feature fusion. 
It improves the DG/UDA performance by +0.9/+1.3 mIoU over the SegFormer decoder (see Tab.~\ref{tab:highlevel_ablation}). Tab.~\ref{tab:decoder_fusion} additionally compares the UDA performance of DAFormer with further decoder designs. In particular, DAFormer outperforms a variant without depthwise separable convolutions and a variant with ISA~\cite{huang2019interlaced} instead of ASPP for feature fusion. This shows that a capable but parameter-effective decoder with an inductive bias of the dilated depthwise separable convolutions is beneficial for good UDA performance.
When the context is only considered for bottleneck features, the UDA performance decreases by -1.3 mIoU, revealing that the context clues from different encoder stages used in DAFormer are more domain-robust.
Tab.~\ref{tab:decoder_fusion} further compares DAFormer to UperNet~\cite{xiao2018unified}, which iteratively upsamples and fuses the features. Even though UperNet achieves the best oracle performance, it is noticeably outperformed by DAFormer on UDA, which confirms that it is necessary to study and design the decoder architecture, along with the encoder architecture, specifically for domain robustness.

\subsection{Domain-Robust Multi-Resolution Training (HRDA)}
\label{sec:exp_resolution_crop_size}
\label{sec:exp_multi_resolution_uda}

\begin{figure}
\centering
\includegraphics[width=\linewidth]{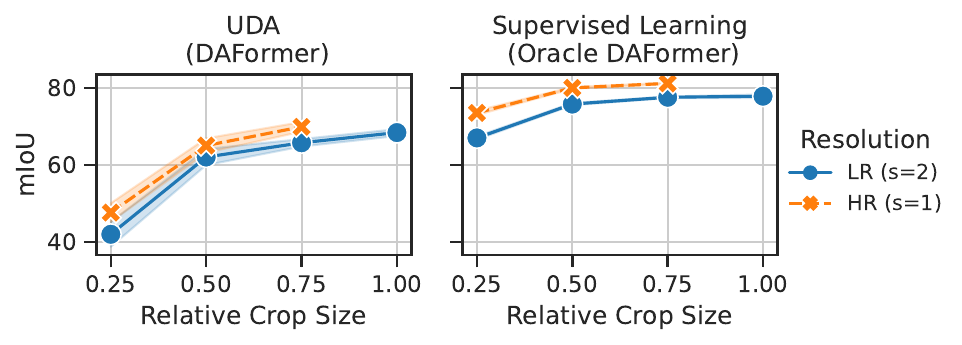}
\caption{Segmentation mIoU over the relative crop size ($h / \frac{H_T}{s}$) for different resolutions and for both DAFormer UDA and the target-supervised oracle. There is no value for $\text{HR}_{1.0}$ due to GPU memory constraints.}
\label{fig:crop_resolution}
\end{figure}

By utilizing the proposed domain-robust multi-resolution framework (HRDA), the DG/UDA performance further improves by +3.9/5.5 mIoU (see Tab.~\ref{tab:highlevel_ablation}). In the following, we provide an in-depth analysis of HRDA for UDA.

\begin{figure}[tb]
\centering
\includegraphics[width=\linewidth]{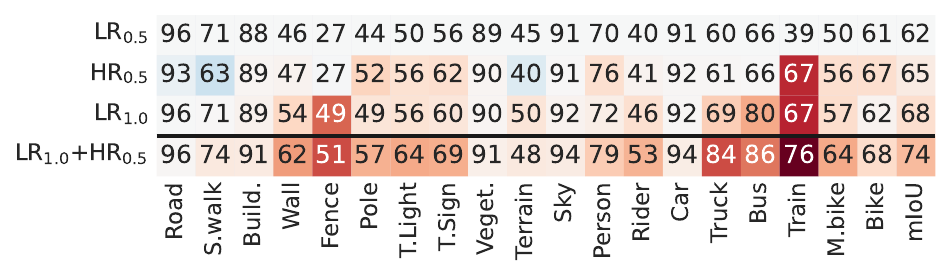}
\caption{
Class-wise IoU for UDA with DAFormer for different crop resolutions $\mathit{XR}$ ($s_\mathit{LR}{=}2, s_\mathit{HR}{=}1$) and relative crop sizes $a{=}h / \frac{H_T}{s_\mathit{XR}}$. Crops are denoted as $\mathit{XR}_a$. The colors indicate the difference to the first row.
}
\label{fig:crop_resolution_heatmap}
\end{figure}

\noindent\textbf{Influence of Resolution and Crop Size:}
As underlying principle of HRDA, we analyze the influence of the resolution and crop size on UDA with a single-crop DAFormer model.
Instead of the crop size in pixels, we use the relative crop size $h / \frac{H_T}{s}$ (with crop height $h$, original image height $H_T$, and downscale factor $s$) to describe how much content from the image is part of the crop independent of its resolution. When referring to small/large crops, we mean the rel. crop size and not the crop size in pixels.
Fig.~\ref{fig:crop_resolution} shows that both an increased resolution and rel. crop size improve the performance for UDA and supervised learning. 
A large crop size is even more important for UDA than for supervised learning, i.e. a 4 times smaller LR crop reduces the performance by 39\% for UDA and by 14\% for supervised training. The larger crop provides more context clues and improves the performance of all classes, especially the ones that are difficult to adapt such as wall, fence, truck, bus, and train (cf. row 1 and 3 in Fig.~\ref{fig:crop_resolution_heatmap}), probably, as the relevant context clues are more domain-invariant~\cite{huang2020contextual, yang2021context}.
A higher input resolution improves the UDA performance by a similar amount as it improves supervised learning. The improvement originates from a higher IoU for small classes such as pole, traffic light, traffic sign, person, motorbike, and bicycle, while some large classes such as road, sidewalk, and terrain have a decreased performance (cf. row 1 and 2 in Fig.~\ref{fig:crop_resolution_heatmap}). This supports that large objects are easier to adapt at LR while small objects are easier to adapt at HR, which can be exploited by the multi-resolution fusion of HRDA.

\noindent\textbf{Combining Multiple Resolutions:} Next, we combine crops from LR and HR using the proposed domain-robust multi-resolution framework. Tab.~\ref{tab:context_crop_size} shows that training with multiple resolutions improves the performance over both LR-only and HR-only training by +3.4 mIoU, which demonstrates that multi-resolution fusion with scale attention improves domain robustness. Based on the observation that large crops are important for UDA, we increase the rel. context crop size to 1.0, which further improves the performance by +5.3 mIoU, demonstrating the effectiveness of the proposed small HR detail and large LR context crops. Fig.~\ref{fig:crop_resolution_heatmap} shows that the multi-resolution training combines the strength of the single-scale training with $\text{HR}_{0.5}$ and $\text{LR}_{1.0}$ as the multi-resolution IoU of each class is better than the best single-scale IoU.

\begin{table}
\centering
\caption{
Combining context and detail crops with HRDA. $\mathit{XR}_a$ denotes crops with resolution $\mathit{
XR}$ ($s_\mathit{LR}{=}2$, $s_\mathit{HR}{=}1$) and relative crop size $a{=}h / \frac{H_T}{s_\mathit{XR}}$.
}
\label{tab:context_crop_size}
\setlength{\tabcolsep}{5pt}
\footnotesize
\begin{tabular}{lll}
\toprule
Context Crop & Detail Crop &           mIoU \\
\midrule
$\text{LR}_{0.5}$ &       -- & 62.1\spm{2.1} \\
-- &   $\text{HR}_{0.5}$ & 65.1\spm{1.9} \\
$\text{LR}_{0.5}$ &   $\text{HR}_{0.5}$ & 68.5\spm{0.6} \\
$\text{LR}_{1.0}$ &   $\text{HR}_{0.5}$ & 73.8\spm{0.3} \\
\bottomrule
\end{tabular}
\end{table}

\begin{table}
\centering
\caption{Comparison of HRDA with naive HR crops that have a comparable GPU memory footprint ($\text{HR}_{0.75}$).}
\label{tab:baselines}
\setlength{\tabcolsep}{3pt}
\footnotesize
\begin{tabular}{llcl}
\toprule
Context Crop & Detail Crop & GPU Memory &            mIoU \\
\midrule
          -- &     $\text{HR}_{0.75}$ &  22.0 GB &  70.0\spm{1.2} \\
     $\text{LR}_{0.75}$ &    $\text{HR}_{0.375}$ &  13.5 GB &  71.3\spm{0.3} \\
      $\text{LR}_{1.0}$ &      $\text{HR}_{0.5}$ &  22.5 GB &  73.8\spm{0.3} \\
\bottomrule
\end{tabular}
\end{table}

\noindent\textbf{Comparison with Naive HR:} We compare HRDA with naive large HR crops ($\text{HR}_{0.75}$), which have a comparable GPU memory footprint as HRDA. This is a very strong baseline, which is already +1.7 mIoU better than DAFormer. Tab.~\ref{tab:baselines} shows that HRDA still outperforms $\text{HR}_{0.75}$ by +3.8 mIoU. Even when reducing the crop size of HRDA to match the size of $\text{HR}_{0.75}$, HRDA is still +1.3 mIoU better while requiring 40\% less GPU memory. This demonstrates that combining LR context crops and HR detail crops performs better than naively increasing the resolution, due to HRDA's capability of capturing long-range context and multi-resolution fusion.

\noindent\textbf{HRDA Component Ablations:}
The components of HRDA are ablated in Tab.~\ref{tab:ablations}.
The most crucial component is the learned scale attention, which improves the mIoU by +3.0 over naively averaging the predictions from both scales.
This shows that it is crucial to learn which scale is best-suited to adapt certain image regions. Generating pseudo-labels with different context views by overlapping detail crops results in a further gain of +0.9 mIoU. Finally, additional supervision of the detail crop ($\lambda_d=0.1$) further provides +1.4 mIoU.

\begin{table}[tb]
\centering
\caption{Detailed ablation of the components of HRDA.}
\label{tab:ablations}
\setlength{\tabcolsep}{3pt}
\footnotesize
\begin{tabular}{llll}
\toprule
Scale Attention & Overlap. Detail Crop & Detail Loss &           mIoU \\
\midrule
Average &                 -- &          -- & 67.5\spm{0.8} \\
Learned &                 -- &          -- & 71.5\spm{0.5} \\
Learned &                \cm &          -- & 72.4\spm{0.1} \\
Learned &                \cm &         \cm & 73.8\spm{0.3} \\
\bottomrule
\end{tabular}
\end{table}

\begin{figure*}[tb]
\centering
\begin{minipage}{0.8\linewidth}
{\footnotesize
\begin{tabularx}{\linewidth}{*{6}{Y}}
Image & LR Pred. & HR Pred. & Scale Attent. & Fused Pred. & G. Truth \\
\end{tabularx}
} %
\includegraphics[width=\linewidth]{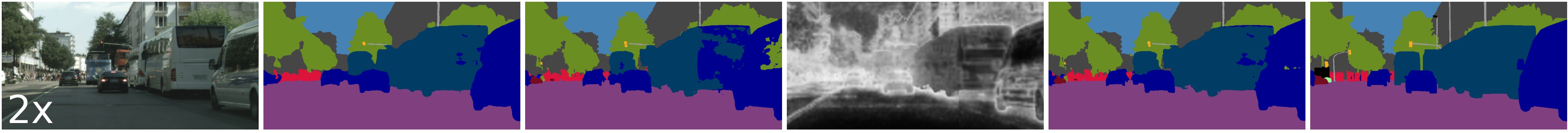}
\includegraphics[width=\linewidth]{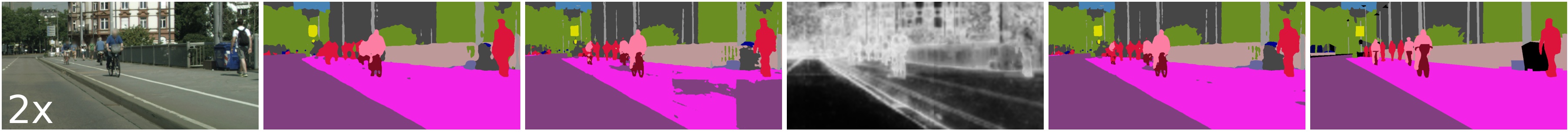}
\includegraphics[width=\linewidth]{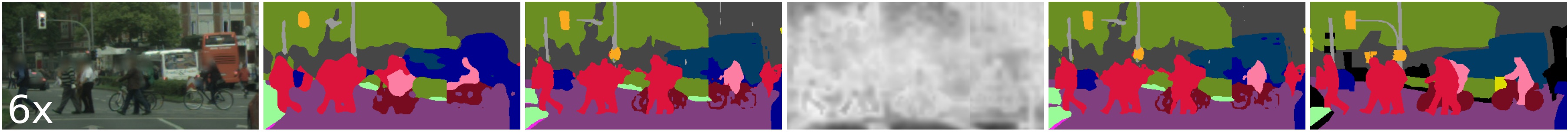}
\end{minipage}
\caption{Examples of the different predictions and the scale attention of HRDA. Large objects are better segmented from LR while small objects are better segmented from HR. The scale attention learns to utilize this pattern for fusing both. The examples are zoomed in (2x or 6x) for better visibility.}
\label{fig:attention_examples}
\end{figure*}

\noindent\textbf{Qualitative Analysis:}
Fig.~\ref{fig:attention_examples} provides representative examples displaying that LR predictions work better for large objects while HR predictions work better for small objects and fine details. The scale attention focuses on LR (black) for large objects and on HR (white) for small objects and segmentation borders, combining the strength of both.

\noindent\textbf{HRDA Oracle:}
\begin{table}[t]
\centering
\caption{
Comparison of UDA, oracle, and relative performance $\mathit{Rel.} = \frac{\mathit{UDA}}{\mathit{Oracle}}$.
}
\label{tab:hrda_oracle}
\begin{tabular}{lccc}
\toprule
         & UDA  & Oracle & Rel.   \\
\midrule
DAFormer & 68.3 \spm{0.5} & 77.6 \spm{0.2} & 88.0\% \\
\;\rotatebox[origin=c]{180}{$\Lsh$} w/ naive HR & 70.0 \spm{1.2} & 81.2 \spm{0.2} & 86.2\% \\
HRDA     & 73.8 \spm{0.3} & 81.6 \spm{0.1} & 90.4\% \\
\bottomrule
\end{tabular}
\end{table}
Training the oracle with naive HR or HRDA increases its mIoU by +3.6/+4.0 (Tab.~\ref{tab:hrda_oracle}). However, the UDA mIoU of HRDA gains +5.5, so that it achieves 90.4\% of the oracle performance, which is 2.4\% more than DAFormer.

\section{Conclusions}
\label{sec:conclusions}

We first presented DAFormer, a network architecture tailored for UDA\&DG. It is based on a Transformer encoder and a context-aware fusion decoder, revealing the potential of Transformers for UDA\&DG. Second, we introduced three training policies to stabilize and regularize UDA\&DG, further enabling the capabilities of DAFormer. Third, we presented HRDA, a multi-resolution approach for UDA\&DG that combines the advantages of small HR detail crops and large LR context crops using a learned scale attention, while maintaining a manageable GPU memory footprint. 

Overall, DAFormer and HRDA jointly achieve unprecedented performances across 5 different benchmarks for synthetic-to-real UDA, day-to-nighttime UDA, clear-to-adverse weather UDA, and synthetic-to-real DG in street scene semantic segmentation. In particular, they reach 73.8 mIoU for UDA on GTA$\rightarrow$Cityscapes, 65.8 mIoU for UDA on Synthia$\rightarrow$Cityscapes, and 55.9 for DG on GTA$\to$Avg.\{Cityscapes,BDD100K,Mapillary\}, which is a respective major gain of +16.3/+10.3/+10.6 mIoU over previous state-of-the-art methods.

\bibliographystyle{IEEEtran}
\bibliography{literature.bib}

\begin{IEEEbiography}[{\includegraphics[width=1in,clip,keepaspectratio]{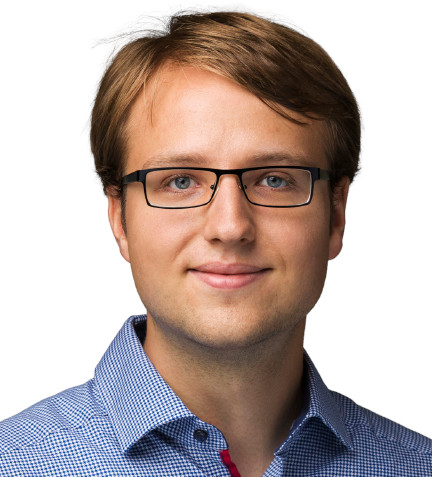}}]{Lukas Hoyer} is a PhD candidate at the Computer Vision Lab at ETH Zurich, Switzerland. His research focuses on label-efficient scene understanding including domain-adaptive, semi-supervised, self-supervised, and multi-task learning for dense prediction tasks such as semantic segmentation and monocular depth estimation. 
He received a MSc in Robotics, Systems and Control from ETH Zurich in 2021 and was honored with the ETH Medal for an outstanding Master's thesis. Before, he studied Computer Systems in Engineering at the University of Magdeburg, Germany and was awarded as best graduate of the computer science department.
\end{IEEEbiography}

\begin{IEEEbiography}[{\includegraphics[width=1in,clip,keepaspectratio]{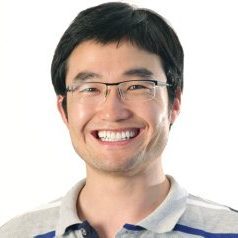}}]{Dengxin Dai} is Director of Research at Huawei Zurich Research Center. Before that, he was a senior group leader at the Max Planck Institute for Informatics and a senior scientist at ETH Zurich. He
received a Ph.D. degree in computer
vision from ETH Zurich in 2016. He is a member of the ELLIS Society. He has organized multiple well-received workshops, is currently an Associate Editor of IJCV, and has been area chair of multiple vision conferences including CVPR, ECCV, and ICRA. He received the Golden Owl Award with ETH Zurich in 2021 for his exceptional teaching and received the German Pattern Recognition Award in 2022 for his outstanding scientific contribution in the area of Scalable and Robust Visual Perception.
\end{IEEEbiography}

\begin{IEEEbiography}[{\includegraphics[width=1in,clip,keepaspectratio]{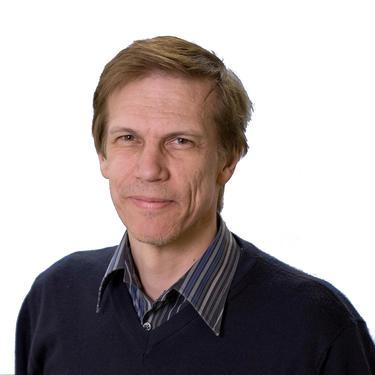}}]{Luc Van Gool}
is currently a full professor for computer vision with ETH Zurich, Switzerland and the
KU Leuven, Belgium. He leads research and
teaches with both places. He has authored more
than 300 papers. He has been a program committee member of several, major computer vision conferences (e.g., program chair ICCV’05, Beijing,
general chair of ICCV’11, Barcelona, and of
ECCV’14, Zurich). His main interests include 3D
reconstruction and modeling, object recognition,
and autonomous driving. He received several best
paper awards (e.g., David Marr Prize ’98, Best Paper CVPR’07). He
received the Koenderink Award, in 2016 and the ‘distinguished researcher’
nomination by the IEEE Computer Society, in 2017. In 2015 he also
received the 5-yearly Excellence Prize by the Flemish Fund for Scientific
Research. He was the holder of an ERC Advanced Grant (VarCity). Currently, he leads computer vision research for autonomous driving in the context of the Toyota TRACE labs in Leuven and at ETH, and has an extensive
collaboration with Huawei on the topic of image and video enhancement.
\end{IEEEbiography}

\vfill

\end{document}